\documentclass{article}

\usepackage{microtype}
\usepackage{graphicx}
\usepackage{booktabs} 
\usepackage[usenames,dvipsnames]{xcolor}

\usepackage[accepted]{icml2021}

\usepackage{hyperref}
\definecolor{mydarkblue}{rgb}{0,0.08,0.45}
\hypersetup{ %
    pdfsubject={Proceedings of the International Conference on Machine Learning 2021},
    colorlinks=true,
    linkcolor=mydarkblue,
    citecolor=mydarkblue,
    filecolor=mydarkblue,
    urlcolor=mydarkblue,
}
\usepackage[nolist]{acronym}
\usepackage{tikz}
\usepackage[font=small, skip=5pt, belowskip=0pt, labelfont={it}]{caption}
\usepackage{subfig}
\usepackage{enumitem}
\usepackage{mathaux}
\usepackage{makecell}

\makeatletter
\newcommand*\bigcdot{\mathpalette\bigcdot@{.5}}
\newcommand*\bigcdot@[2]{\mathbin{\vcenter{\hbox{\scalebox{#2}{$\m@th#1\bullet$}}}}}
\makeatother

\definecolor{538blue}{RGB}{48,162,218}
\definecolor{538red}{RGB}{252,79,48}
\definecolor{538yellow}{RGB}{229,174,56}
\definecolor{538green}{RGB}{109,144,79}
\definecolor{538purple}{RGB}{129, 15, 124}
\definecolor{538grey}{RGB}{139,139,139}
\definecolor{538magenta}{RGB}{200,100,200}

\DeclareRobustCommand\legend[1]{%
  \tikz\draw[#1] (0,0) (0,\the\dimexpr\fontdimen22\textfont2\relax)
  -- (8pt,\the\dimexpr\fontdimen22\textfont2\relax);%
}

\newcommand{\catgames}{54}
\newcommand{\dqngames}{54}

\newacro{drl}[DRL]{Deep Reinforcement Learning}
\newacro{dqn}[DQN]{Deep Q-Networks}
\newacro{rl}[RL]{Reinforcement Learning}
\newacro{sn}[SN]{Spectral Normalisation}
\newacro{sr}[SR]{Spectral Norm Regularisation}
\newacro{sl}[SL]{Supervised Learning}
\newacro{bn}[BN]{Batch Normalisation}
\newacro{gp}[GP]{Gradient Penalty}
\newacro{wn}[WN]{Weight Normalisation}
\newacro{wd}[WD]{Weight Decay}
\newacro{c51}[C51]{Categorical DQN}
\newacro{dqn}[DQN]{Deep Q-Networks}
\newacro{ale}[ALE]{Arcade Learning Environment}
\newacro{td}[TD]{Temporal-difference learning}
\newacro{gan}[GAN]{Generative Adversarial Network}
\newacro{gans}[GANs]{Generative Adversarial Networks}  
\newacro{mdp}[MDP]{Markov Decision Process}
\newacro{mdps}[MDPs]{Markov Decision Processes}  
\newacro{mlp}[MLP]{Multi-Layer Perceptron}  
\newcommand{\rainbow}{\textsc{Rainbow}}

\icmltitlerunning{Spectral Normalization for Deep Reinforcement Learning: an Optimisation Perspective}

\begin{document}

\twocolumn[
    \icmltitle{Spectral Normalisation for Deep Reinforcement Learning:\\ An Optimisation Perspective}
    \icmlsetsymbol{equal}{*}
    \begin{icmlauthorlist}
        \icmlauthor{Florin Gogianu}{equal,bit,utc}
        \icmlauthor{Tudor Berariu}{equal,icl}
        \icmlauthor{Mihaela Rosca}{dm,ucl}
        \icmlauthor{Claudia Clopath}{icl,dm}
        \icmlauthor{Lucian Busoniu}{utc}
        \icmlauthor{Razvan Pascanu}{dm}
    \end{icmlauthorlist}
    \icmlaffiliation{bit}{Bitdefender, Bucharest, Romania}
    \icmlaffiliation{icl}{Imperial College London, Department of Bioengineering, London, UK}
    \icmlaffiliation{dm}{DeepMind, London, UK}
    \icmlaffiliation{ucl}{Centre for Artificial Intelligence, University College London, London, UK}
    \icmlaffiliation{utc}{Department of Automation, Technical University of Cluj-Napoca, Romania}
    \icmlcorrespondingauthor{Tudor Berariu}{tudor.berariu@gmail.com}
    \icmlkeywords{Machine Learning, ICML, Reinforcement Learning, Atari, Adam, Optimisation}
    \vskip 0.3in
]

\printAffiliationsAndNotice{\icmlEqualContribution} 

\begin{abstract}
    Most of the recent deep reinforcement learning advances take an RL-centric perspective and focus on refinements of the training objective.
    We diverge from this view and show we can recover the performance of these developments not by changing the objective, but by regularising the value-function estimator.
    Constraining the Lipschitz constant of a single layer using spectral normalisation is sufficient to elevate the performance of a Categorical-DQN agent to that of a more elaborated \rainbow{} agent on the challenging Atari domain.
    We conduct ablation studies to disentangle the various effects normalisation has on the learning dynamics and show that is sufficient to  modulate the parameter updates to  recover most of the performance of spectral normalisation.
    These findings hint towards the need to also focus on the neural component and its learning dynamics to tackle the peculiarities of Deep Reinforcement Learning. 
\end{abstract}

\section{Introduction}
\label{sec:introduction}

Deep Reinforcement Learning has made considerable strides in recent years, scaling up to complex games as Go and Starcraft.
However, besides relying on neural network function approximation, by and
 large the community has taken an RL-centric stance, with a plethora of approaches from prioritised replay~\citep{Schaul2016PrioritizedER} to improving exploration~\citep{Fortunato2018NoisyNF, Osband2016DeepEV}.
Many of these advances have been collated in the \rainbow{} agent \citep{hessel2018RainbowCI}, a strong single-threaded \ac{drl} algorithm on the Atari \ac{ale} benchmark \citep{bellemare2013arcade}.
We take an orthogonal approach by focusing instead on the neural network learning dynamics.
Most of modern deep learning assumes the optimisation of an objective on a finite and fixed data-set where i.i.d. sampling is possible.
These assumptions are rarely satisfied in \ac{drl}, where data is non-stationary, the observed samples tend to be highly correlated, and sometimes the updates do not form a gradient vector field~\citep{czarnecki2019distilling,bengio2020InterferenceAG}. 
These shortcomings are often reflected in practice, as attested by multiple reports of narrow good hyper-parameter ranges for optimisation \citep{Henderson2018WhereDM,Henderson2018DeepRL}, under-generalisation in \acf{td} \citep{bengio2020InterferenceAG}, and, as opposed to the supervised setting, ineffective regularisation where \ac{bn}, dropout or L2 hurt performance \citep{bhatt2019crossnorm,liu2019regularization,Cobbe2019QuantifyingGI}.
We emphasise the importance of studying the learning dynamics of neural networks in \ac{drl} by showing the gains such studies can provide. 

\begin{figure}[t]
    \centering
    \includegraphics[width=\columnwidth]{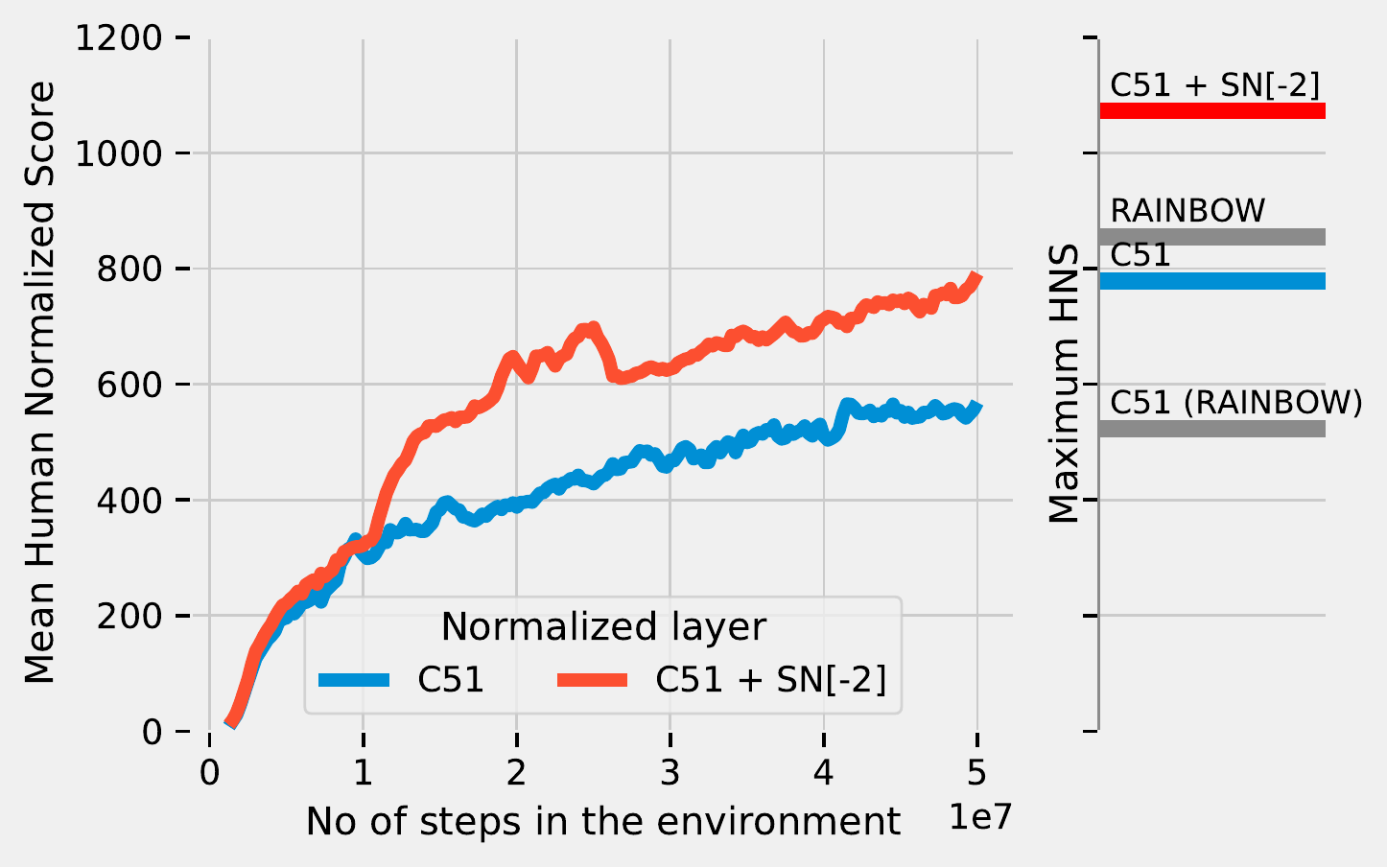}
    \caption[Baseline vs Spectral Norm]{
        \textbf{Optimisation rivals algorithmic improvements.} 
        \newline 
        \textbf{Left:} \underline{Average} Human Normalised Score (HNS) per time-step for C51 with \acf{sn}.
        \textbf{Right:} Comparison of \underline{Maximum} HNS  reported in the literature (same as in Table \ref{tbl:hns}). Average over \catgames{} \acf{ale} games.}
    \label{fig:atari_c51_hns}
    \vspace{-1em}
\end{figure}

In our paper we explore \acf{sn}, one such technique originating in the \ac{gans} literature \citep{miyato2018spectral}, another field having to deal with non-stationarity and challenging optimisation dynamics.
\ac{sn} was primarily used for its \emph{regularisation} effect of controlling the smoothness of the function.
In \citep{farnia2018generalizable} authors rely on \ac{sn} as a defense against adversarial examples, \citep{anil2019sorting} for improving estimation of Wasserstein distance and 
\citep{yu2020mopo} to regularise model-based \ac{rl}.

However \ac{sn} also decouples the norm of the weight vector from its direction, therefore affecting \emph{optimisation} in subtle ways, especially in the case of adaptive, momentum based methods \citep{rosca2020case}.

Our main contributions in this paper are:

\textbf{Efficiency of \ac{sn} in \ac{drl}.}
We show that \ac{sn} can lead to \emph{performance gains that rivals other algorithmic improvements} such as those collated in the \rainbow{} agent (Fig.~\ref{fig:atari_c51_hns}). We additionally provide code to ensure reproducibility.\footnote{\url{https://github.com/floringogianu/snrl}}
    
\textbf{Optimisation effect of \ac{sn}.} We investigate the role of \ac{sn} and argue that \emph{its effect is primarily an optimisation one}.
Although we identify a small correlation between smoothness and performance in our careful ablation studies, we argue the effect of smoothness is secondary.
In particular, we can mimic the impact of \ac{sn} on the parameter updates without changing the function represented by the neural network, hence without smoothness constraints, recovering the performance of \ac{sn}. 
 
\textbf{Reducing hyper-parameter sensitivity.} We show through large scale experiments that \ac{sn} can extend Adam's range of usable hyper-parameters.

\section{Background}
\label{sec:background}
\subsection{Value based deep reinforcement learning}
\label{sub:background--drl}

An \ac{rl} problem is usually described as a \ac{mdp} $\mathcal{M} = (\mathcal{S}, \calA, \trans, r, \rhoinit, \gamma)$ consisting of state space $\mathcal{S}$, the set of all possible actions $\calA$, a transition distribution $\trans(s'|s,a)$ and a reward function $r: \calS\times \calA \rightarrow \mathbb{R}$. The goal of an agent is to find a policy $\pi$ that maximises the expected sum of discounted rewards: $Q^{\pi}(s, a) = \mathbb{E}_{s_0\sim\rhoinit,a_t~\sim\pi\left(s_t\right)}\left[\sum_{t=0}\gamma^t r(s_t, a_t)\right]$.
An important class of algorithms rely on state-action values to find $\pi$. 
These approaches focus on minimising the \emph{temporal difference} (TD) error to learn the $Q$-function and define $\pi$ greedily with respect to it.
We are particularly interested in scenarios where $Q$ is approximated by a neural network. 
The prototypical algorithm is DQN~\citep{mnih2015human}, where $Q$ is learned by minimising:
\begin{equation}
\calL(\theta) = \underset{\scriptsize(s, a, r, s^{\prime}) \sim \calD}{\mathbb{E}}\Big(Q_{\theta}(s, a) -  \big(r + \gamma \max_{a^{\prime}}Q_{\theta^{\prime}}(s^{\prime}, a^{\prime})\big)\Big)^2
\label{eqn:dqn}
\end{equation}
In TD error (Eq.~\ref{eqn:dqn}), the bootstrapped term uses held-back parameters $\theta^{\prime}$ to bring extra stability. During training random batches of transitions are sampled from the last million experiences which are kept in a replay buffer $\mathcal{D}$. This mitigates, but does not enforce the i.i.d. and stationarity assumption required by stochastic gradient descent. 

Since DQN first reported human-level results on the ALE Atari environment, several refinements of the algorithm were proposed. A remarkable line of research proposes to model the distribution of the state-action value rather than just its mean. For example, C51 \citep{bellemare2017distributional} predicts a categorical distribution of expected returns. \rainbow{} \citep{hessel2018RainbowCI} integrates several advances on top of C51: prioritised experience replay for improved data sampling \citep{Schaul2016PrioritizedER}, n-step returns for lower variance bootstrap targets, double Q-learning for unbiasing estimates \citep{van2016deep}, disambiguation of state-action value estimates \citep{wang2016dueling} and noisy nets for improved exploration \citep{Fortunato2018NoisyNF}.

An often overlooked aspect is that the emergence of RMSProp \cite{Tieleman2012} and Adam \cite{kingma2014adam} adaptive optimisation algorithms is closely linked with the recent success of \ac{drl}.

\subsection{Spectral Normalisation}
\label{sub:background--sn}

\acf{sn} is an approach for controlling the Lipschitz constant of certain families of parametric functions such as linear operators.
While it can be defined more generally, for our purpose we will consider a function to be Lipschitz continuous in the $\ell_2$ norm if $\lVert f(\bm{x}_1) - f(\bm{x}_2) \rVert_2  \leq  k \lVert \bm{x}_1 - \bm{x}_2 \rVert_2$ and we call $k$ the \emph{Lipschitz constant} of the function.
A linear map $\bm{y} = \weight{}\bm{x}$ is 1-Lipschitz ($k=1$) if $\lVert \weight{}\bm{x}\rVert_2 \leq \lVert\bm{x}\rVert_2$ for any $\bm{x}$.
This makes it apparent that normalising the spectral radius (largest singular value) of $\weight{}$ enforces the 1-Lipschitz constraint on the linear map $\weight[sn]{} = \weight{} / \left\lVert \weight{} \right\rVert_2 = \rho^{-1} \weight{}$. This holds for convolutions as they are linear operators. A function is $k$-Lipschitz if the largest singular value of $\lVert \mathbf{J}_{\mathbf{xy}} \rVert$ is bounded by $k$.
We make use of this relation to characterise the local smoothness of the learned neural networks in Sec.~\ref{sub:analysis--smoothness}.

The Lipschitz constant of a composition of two functions, $f_1$ with Lipschitz constant $k_1$ and $f_2$ with constant $k_2$, will be bounded by the product $k_1 \cdot k_2$.
This implies that the Lipschitz constant of a neural network can be bounded by setting the Lipschitz constant of each layer.
For a complete overview over the Lipschitz constant of various layers, pooling and common activation functions including ReLU we refer to \citep{anil2019sorting,gouk2020regularisation}.

\section{Methods}
\label{sec:methods}

Computing the spectral radius at each training step would be prohibitive, therefore we approximate it with one step of power iteration \citep{GOLUB200035} at every forward pass. The algorithm is sketched below with $\mathbf{u}$ and $\mathbf{v}$ being the right, and left singural vectors.
\begin{align*}
    \mathbf{v} &\leftarrow \weight{}\mathbf{u}^{(t-1)}; & \alpha &\leftarrow \lVert \mathbf{v}\rVert; & \mathbf{v}^{(t)} &\leftarrow \alpha^{-1} \mathbf{v} \\
    \mathbf{u} &\leftarrow \trn{\weight{}}\mathbf{v}^{(t)}; & \rho &\leftarrow \lVert \mathbf{u}\rVert; & \mathbf{u}^{(t)} &\leftarrow \rho^{-1} \mathbf{u}
\end{align*}
We then use the spectral radius estimation to perform a hard projection of the parameters:
$\weight[sn]{} = \weight{} / \max (\lambda, \; \rho)$.
In all our experiments $\lambda=1$ if not specified otherwise.
For convolutional layers we adapt the procedure in \citep{gouk2020regularisation} and the two matrix-vector multiplications are replaced by convolutional and transposed convolutional operations.

\subsection{Relaxations of Spectral Normalisation}
\label{sub:methods--relaxations_of_sn}

Our first experiments with \ac{sn} quickly highlighted several interesting observations.
First, constraining all layers has a negative impact on the agent's performance.
This is not surprising, a priori there is no reason for the true $Q$-functions to be smooth.
Actions can often lead to drastic changes in the score, where taking a single action at time $t$ forfeits the game while the same action at previous step might carry no significance, as 
for example in \textit{Pong} when the paddle is close to the ball or in \textit{MS-PacMan} when a ghost is close to the agent.
This require very different values for close-by states, making the optimal $Q$ function non-smooth. 
%

These initial findings made it apparent that controlling the amount of regularisation 
is required.
%
%
To address similar issues, \citep{gouk2020regularisation} proposes changing the projection rule to $\weight[sn]{i} = \weight{i} / \max(\lambda_i, \lVert \weight{i} \rVert_2)$ and conduct a hyper-parameter search for the layer specific constants $\lambda_i$.

In contrast, \emph{we apply the normalisation only on a few layers of the network}.
We found it best to apply normalisation to the layers with the largest number of weights (Figs.~\ref{fig:atari_dqn_hns}, \ref{fig:minatar_SN_depth_vs_width_TWO}). 
For value functions networks in \ac{drl}, such as the classic \ac{dqn} architecture, this tends to be the layer before the output layer.
%
%
%
We further motivate and discuss our design decisions in Sec.~\ref{sub:analysis--smoothness_too_much}.

Because of the bias towards normalising layers deeper in the network we will refer to the index of the layer being normalised in a notation akin to \texttt{Python} list indexing syntax.
For layers of any depth \texttt{SN[-1]} will refer to the output layer and \texttt{SN[-2]} to the one before it, while \texttt{SN[-2,-3]} refers to both layers being normalised together.


\section{Spectral Normalisation on the Atari suite}
\label{sec:atari}

Our intuition is that \ac{sn} plays a dual role: one of smoothing the function and one of altering the optimisation dynamics of the neural network.
This dual role has already been hinted to in other settings with non-stationary effects, such as \ac{gans} \citep{rosca2020case}.
\ac{drl} is also characterised by problematic data shifts during training, therefore we seek to find out whether \ac{sn} has a positive effect for RL-agents as well, and isolate its effect to \ac{sn}'s ability to either smooth the function or to affect the optimisation process. 

\begin{figure}[t]
    \centering
    \includegraphics[width=\columnwidth]{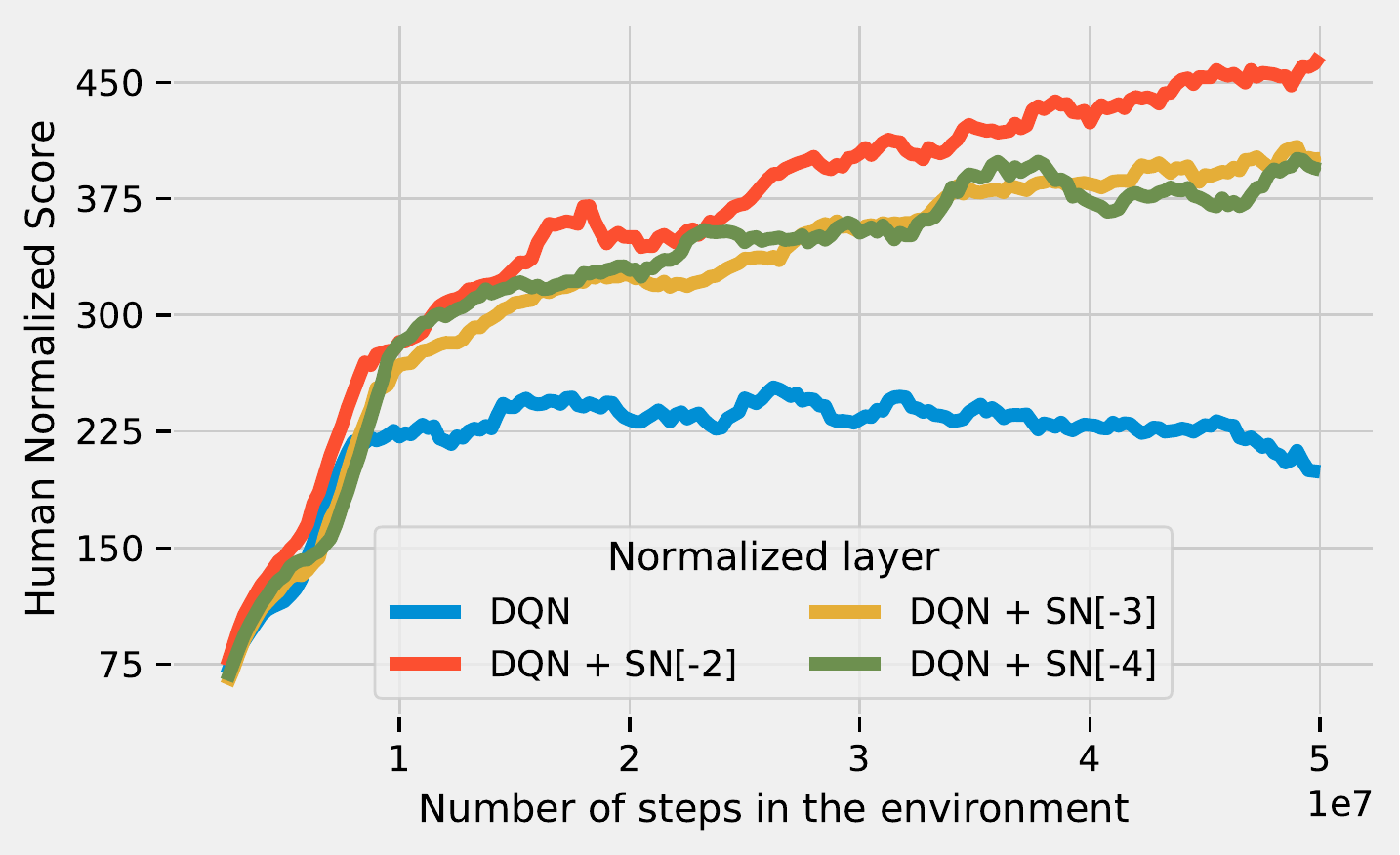}
    \caption[Baseline vs Spectral Norm]{
        Human Normalised Score for a DQN-Adam baseline with \ac{sn} applied on three different layers.
        Average over \dqngames{} Atari games.}
    \label{fig:atari_dqn_hns}
\end{figure}

To this end, we evaluate \ac{sn} on the \acf{ale} \citep{bellemare2013arcade}. This collection of Atari games is varied and complex enough to ensure the generality of our claims and observations.
Normalising only the penultimate layer (\texttt{SN[-2]}) while keeping everything else fixed enhances the performance of \acf{c51} beyond that of \rainbow (Fig.~\ref{fig:atari_c51_hns}, Table~\ref{tbl:hns}) an agent that aggregates the advantages of many other RL advances on top of \ac{c51}.

\begin{table}[t]
    \vskip 0.15in
    \begin{center}
        \begin{small}
            \begin{tabular}{lcccr}
                \toprule
                \textsc{Agent}                                  & \textsc{Mean}             & \textsc{Median} \\
                \midrule
                \textsc{DQN} \cite{wang2016dueling}             & \textsc{216.84}           & \textsc{78.37}  \\
                \textsc{DQN-Adam$^{\ast}$}                      & \textsc{358.45}           & \textsc{119.45} \\
                \textsc{DQN-Adam SN[-2]}                        & \textsc{\textbf{719.95}}  & \textsc{\textbf{178.18}} \\
                \hline%
                \rule{0pt}{2ex}%
                \textsc{C51} \cite{hessel2018RainbowCI}         & \textsc{523.06}           & \textsc{146.73} \\
                \textsc{C51} \cite{bellemare2017distributional} & \textsc{633.49}           & \textsc{174.84} \\
                \textsc{C51$^{\ast}$}                           & \textsc{778.68}           & \textsc{182.26} \\
                \textsc{Rainbow} \cite{hessel2018RainbowCI}     & \textsc{855.11}           & \textsc{227.05} \\
                \textsc{C51 SN[-2]}                             & \textsc{\textbf{1073.18}} & \textsc{\textbf{248.45}} \\
                \bottomrule
            \end{tabular}
        \end{small}
    \end{center}
    \caption{
        Mean and median Human Normalised Score on \dqngames{} Atari games with random starts evaluation.
        References indicate the sources for the scores for each algorithm.
        We mark our own implementations of the baseline with $\ast$. Our agents are evaluated with the protocol in \cite{hessel2018RainbowCI}.
    }
    \label{tbl:hns}
\end{table}

\begin{figure*}[t]%
     \centering%
     \subfloat{
         \includegraphics[width=0.725\textwidth]{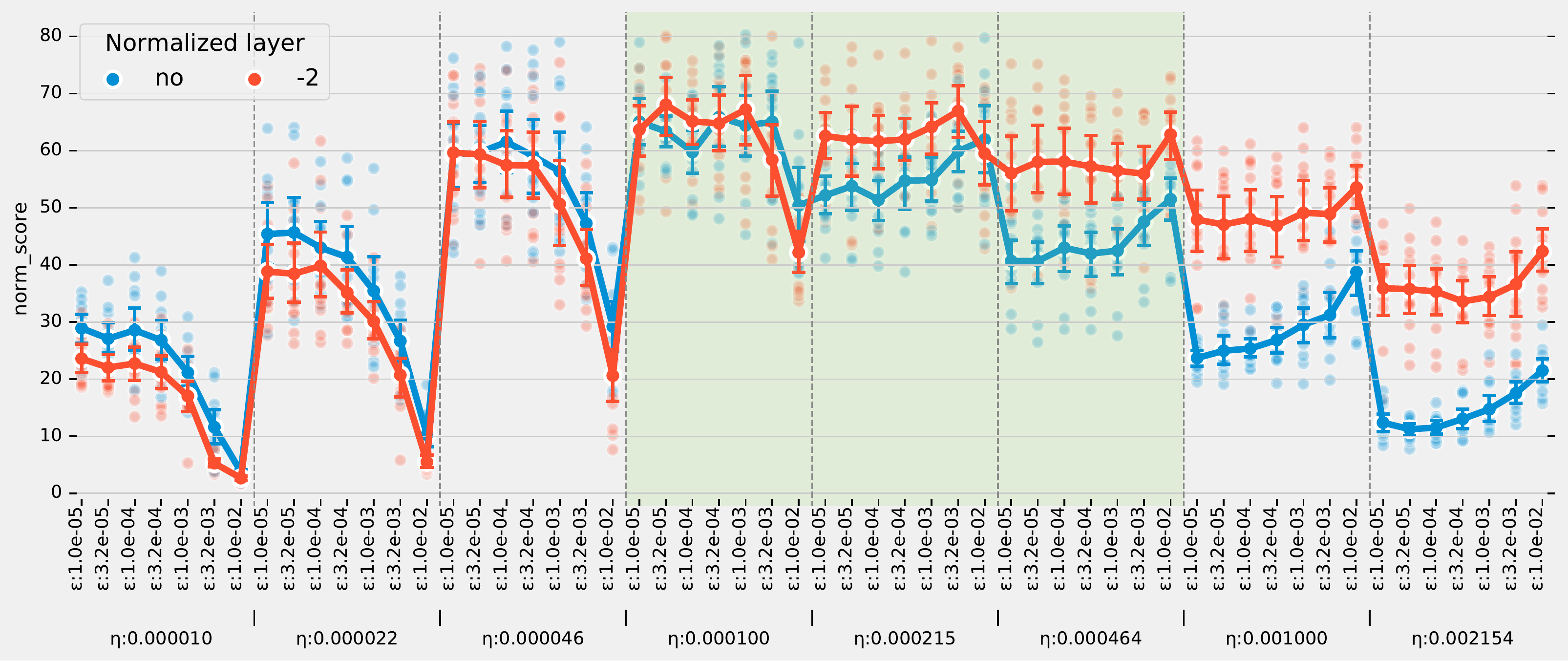}%
     }
     \subfloat{
         \includegraphics[width=0.2675\textwidth]{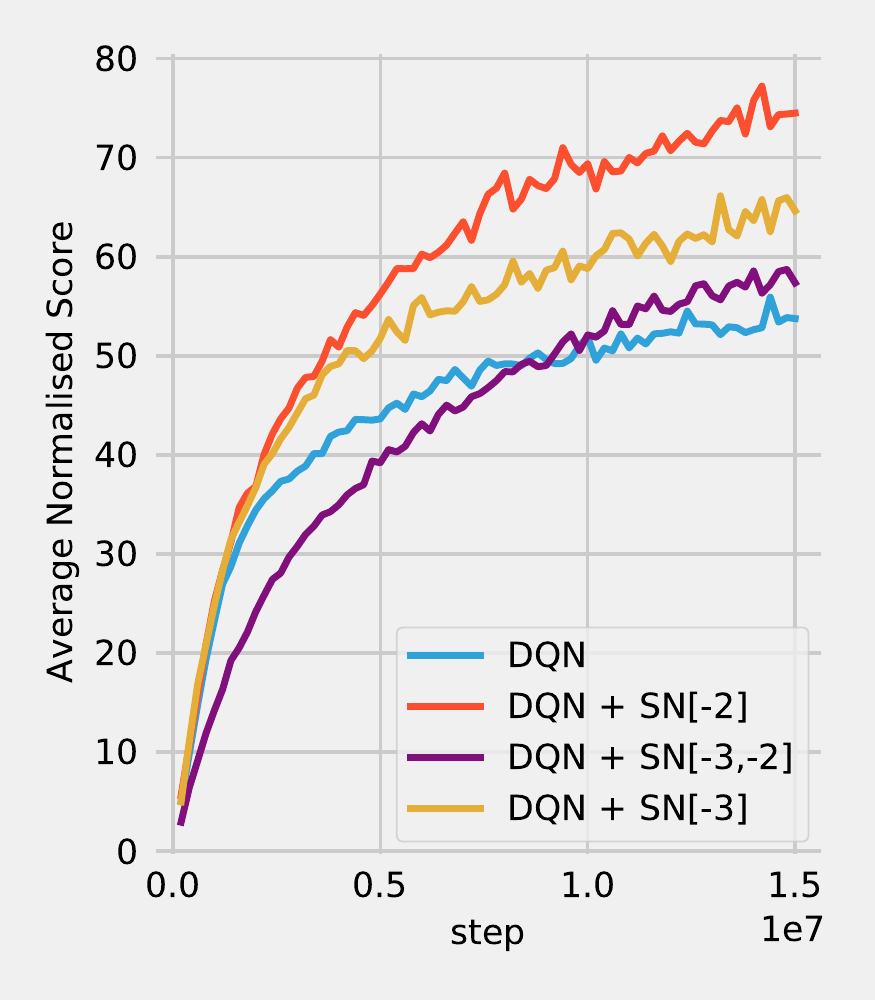}
     }
    \caption{
        \textbf{Left: SN decreases hyper-parameter sensitivity in Adam.}
        Each $\{\eta, \epsilon\}$ combination is used to train 6 models of different depth and width on four MinAtar games, two seeds each.
        Dots represent the average maximum normalised
        score over the four games.
        \textbf{Right: \ac{sn} networks are more adaptable.}
        On a $15$M steps training run the normalised agents continue to learn, adapting to increasingly difficult dynamics while the baseline plateaus.
        Each line is an average of 10 seeds over four MinAtar games and four different model sizes.
        }
    \label{fig:robustness_and_continual_adaptability}
\end{figure*}
\ac{sn} also demonstrates strong performance in the case of non-distributional value-based methods.
Results on smaller environments suggested that \ac{sn} is less effective in conjunction with RMSProp therefore we train a DQN agent using the Adam optimiser instead of RMSProp.
We use the optimiser settings and other hyper-parameters found in Dopamine \citep{castro2018dopamine} as detailed in appendix~\ref{sec:app-atari--dqn_adam}.
We keep the rest of the training and evaluation protocol similar to that of the other algorithms we compare with.
Note that with this change we are not weakening the baseline.
We show that applying \ac{sn} to any of the hidden layers significantly improves upon the DQN baseline (Fig.~\ref{fig:atari_dqn_hns}).
Moreover, the performance closes in on our own implementation of \ac{c51} further suggesting that performance gains coming from optimisation and regularisation advances may be comparable to those achieved by RL-centric methods.
We consolidate all these results in Table \ref{tbl:hns} on a subset of \dqngames{} games%
\footnote{Only \dqngames{} out of 57 because \emph{Surround} ROM is not part of the Atari-Py library we are using, \emph{Defender} crashes on the current version of the library and \emph{Video Pinball} dominates the score due to low human scores and induces noise in the comparisons.}
.
We note that we did not fine-tune our normalised variants but rather reused the same hyper-parameters as for the baseline. 
These hyper-parameters are the result of careful tuning reported in the literature and are detailed in appendix~\ref{sec:app-atari}. We also report results for DQN trained with RMSprop in Sec.~\ref{sec:app-dqn-nature}.

The mean and median Human Normalised Score have been critiqued in the past \citep{machado2018revisiting,toromanoff2019deep} because they can be dominated by some large normalised scores.
However we notice that with very few exceptions ($3/54$ for DQN and $11/54$ games for C51, Fig.~\ref{sec:app-atari}) normalising at most one of the network's layers will not degrade the performance compared to the baseline but instead it will improve upon it, often substantially.

\section{Analysis of results}
\label{sec:analysis}

In order to understand the strong performance of \ac{sn} in Atari we proceed to answer several questions employing both large-scale experiments and technical arguments: How is \ac{sn} interfering with optimisation? Is smoothness a factor in the performance we are observing? And are other regularisation methods leading to similar empirical gains?

\begin{figure}[ht]
    \centering
    \includegraphics[width=\columnwidth]{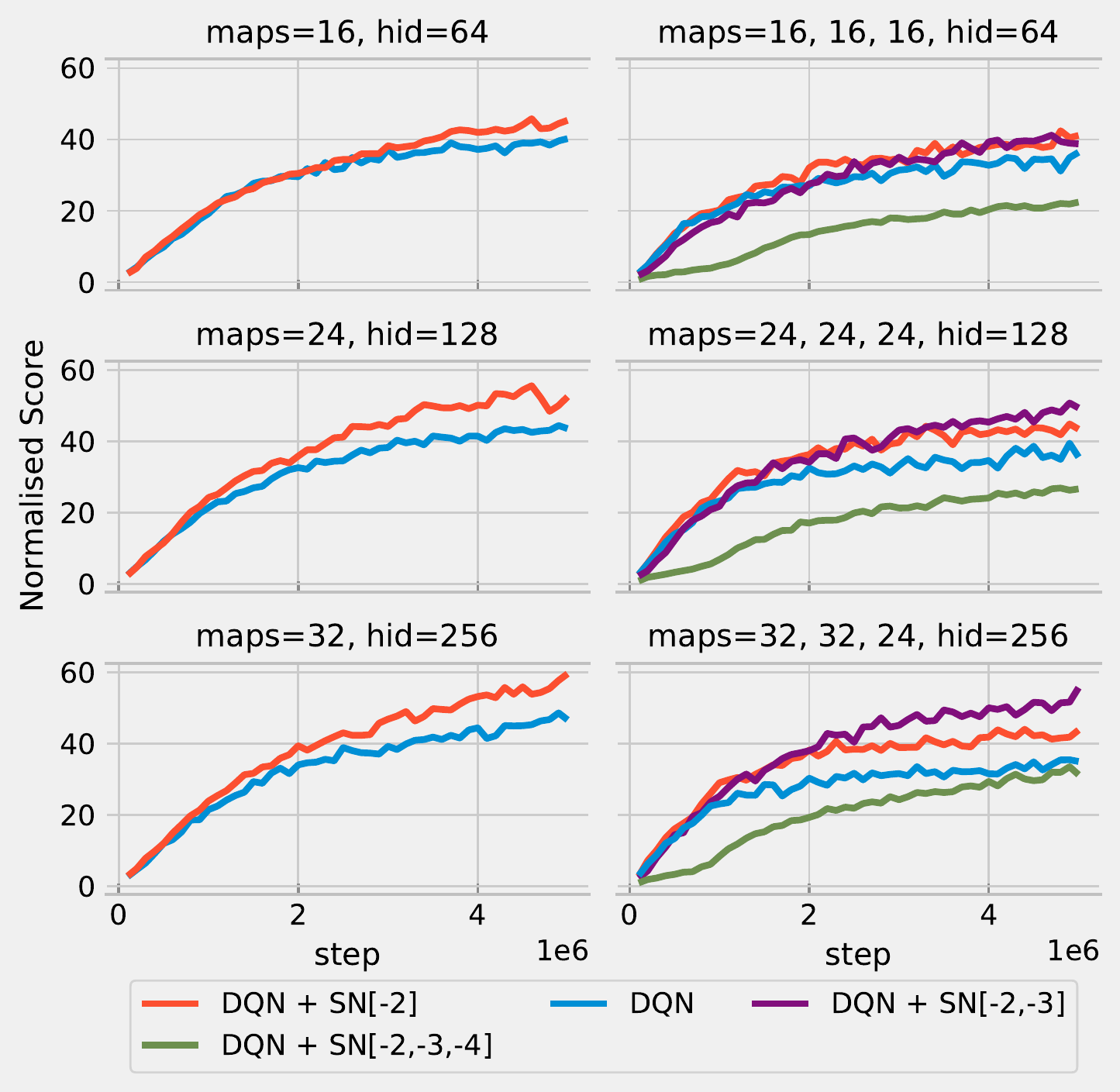}
    \caption{%
    \textbf{\ac{sn} shows gains for all model sizes.}
    We note two regimes of the DQN baseline (\legend{solid,line width=1.5pt,538blue}): for shallow models performance increases with the width of the model; for deeper models performance stagnates with increasing depth and width.
    In both regimes applying SN on individual (\legend{solid,line width=1.5pt,538red}) or multiple (\legend{solid,line width=1.5pt,538purple}) layers improves on DQN suggesting a regularisation effect we could not reproduce with other methods (Fig. \ref{app-fig:minatar_REG_depth_vs_width}).
    Normalising too many layers (\legend{solid,line width=1.5pt,538green}) decreases the capacity and is detrimental to learning.
    Normalised scores of 4 games $\times$ 10 seeds. Details in Fig. \ref{app-fig:minatar_SN_depth_vs_width_FULL}.}
    \label{fig:minatar_SN_depth_vs_width_TWO}
\end{figure}

Since the large scale study we require would have been prohibitive in \ac{ale}, all the experiments in this section if not mentioned otherwise are using for evaluation the MinAtar environment \citep{young19minatar} which reproduces five games from \ac{ale} complete with their non-stationary dynamics but without the visual complexity.
MinAtar was found to be well-suited for reproducing the ablations of recent \ac{drl} contributions originally made on \ac{ale} \citep{ObandoCeron2020RevisitingRP} and therefore we adopt it for our purposes.

Most of the MinAtar results we report are averages of normalised scores over four games.
All agents employ an architecture with at least one convolutional layer and two final linear layers.
When varying the depth of the architecture we only adjust the number of convolutional layers.
When varying the width we increase the number of both feature maps and units in the linear layers.

\subsection{Spectral Normalisation decreases hyper-parameter sensitivity}
\label{sub:results--robustness}

We characterise empirically the impact \ac{sn} has on an agent's performance with a wide variety of architectures and optimisation settings.
In particular, we consider six models of different depths and widths, many learning rates and epsilon values for both Adam and RMSProp with and without \ac{sn} or layer \texttt{[-2]} or layers \texttt{[-2,-3]} for four different games, two seeds for each configuration, resulting in $14000$ DQN agents. 
Fig. \ref{fig:robustness_and_continual_adaptability} reveals several interesting observations:

\begin{enumerate}[noitemsep,topsep=0pt]
    \item \ac{sn} almost never degrades the performance of the baseline therefore it can be safely used in existing agents.

    \item\ac{sn} extends the usable range of Adam's hyper-parameters. Assuming ideal hyper-parameters for the un-normalised network, which typically reside in a very narrow hyper-parameter subspace,  
    applying \ac{sn} does not harm performance. However the normalised model performs well within a significantly larger range, highlighted by the green band in Fig.~\ref{fig:robustness_and_continual_adaptability}, while the baseline degrades dramatically.

    \item\ac{sn} allows for increased adaptability to changing dynamics in the environment.
    MinAtar share with \ac{ale} the property of increasing in difficulty as the policy improves.
    On a 15M training run (Fig. \ref{fig:robustness_and_continual_adaptability}, right) \ac{sn} demonstrates it can continue to learn better policies while the baseline plateaus. This finding is also supported on Atari (Fig.~\ref{fig:atari_dqn_hns}).
\end{enumerate}

\subsection{Smoothness and performance are weakly correlated}
\label{sub:analysis--smoothness}

The Lipschitz constant of a neural network can be at most the product of the Lipschitz constants of its layers.
Since in this work we normalise only a subset of layers, we naturally ask whether it is enough to produce smoother functions. Also, we seek to answer whether performance correlates with the smoothness of the function.

While the exact computation of $k$ for a network is NP-hard \citep{virmaux2018lipschitz} we can compute a \emph{local} smoothness measure with respect to the data distribution.
To approximate the smoothness of our learned value functions we trained agents equipped with twelve different model sizes.
For each game-architecture combination we trained a DQN baseline and normalised various layers, 10 seeds each, resulting in 2400 trained agents.
We then used the best checkpoint for each combination to sample 100k steps in the game and compute for each state $\mathbf{s}$ the norm of the gradient of the state-action value with respect to the state and we report the largest norm encountered.

\begin{figure}[t]
    \centering
    \includegraphics[width=\columnwidth]{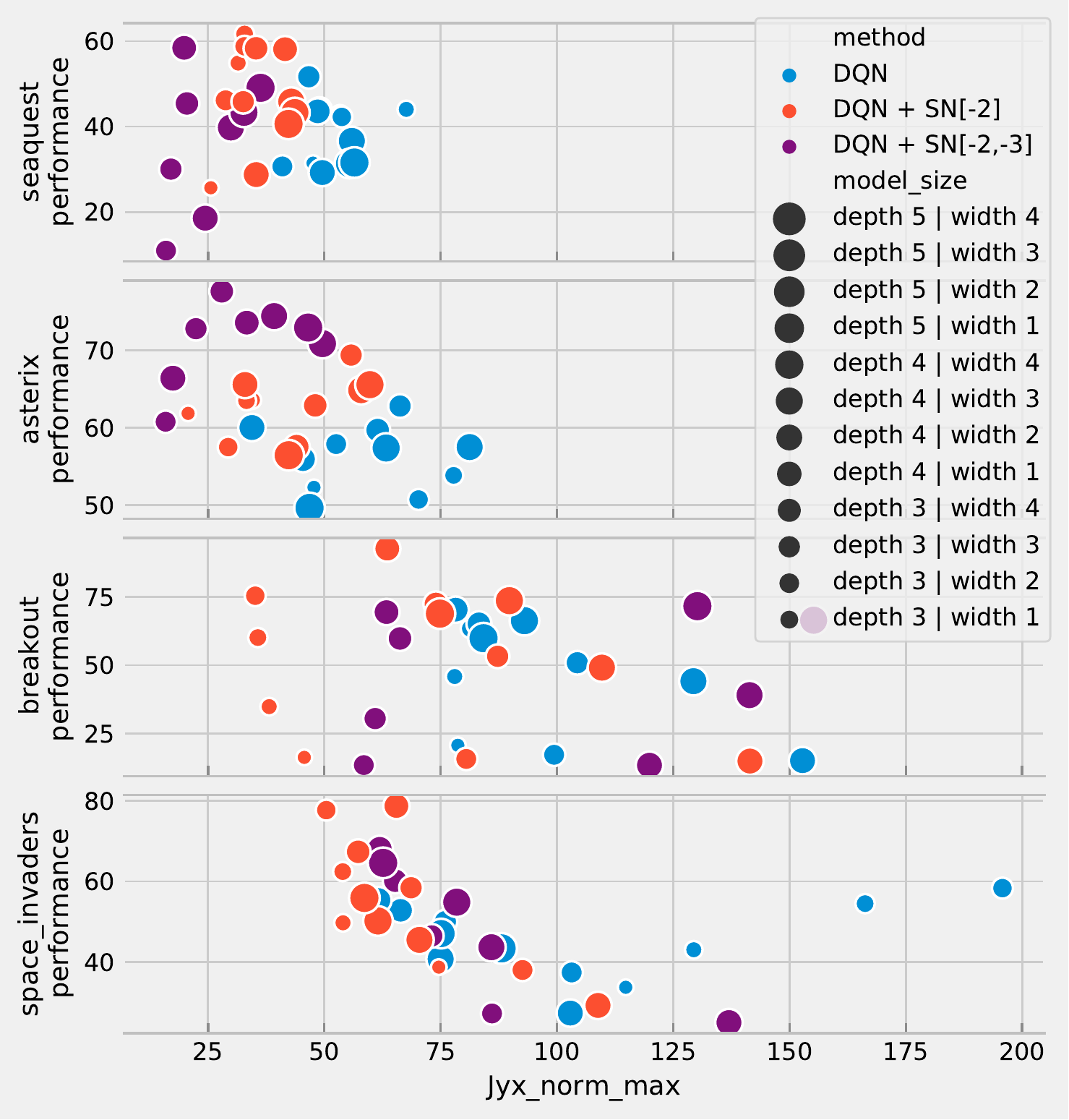}
    \caption[Baseline vs Spectral Norm]{
        \textbf{\ac{sn} does not consistently produce smoother networks.} 
        Often normalising a subset of the network's layers makes the network less smooth than the baseline while performance improves still.
        MinAtar games, average over 10 seeds, for completeness refer to Fig. \ref{app-fig:minatar_smoothness_vs_perf_vs_size_FULL}}
    \label{fig:minatar_smoothness_vs_perf_vs_size}
\end{figure}

We note that while a slight correlation can be measured using Spearman's rank, \ac{sn} does not \emph{generally} produce smoother networks.
The relation between performance and smoothness is not consistent across MDPs, ranging from strong dependence in Space Invaders to no correlation in Breakout (see Fig.~\ref{fig:minatar_smoothness_vs_perf_vs_size}).
The baseline of a three-layer network exhibits a higher empirical norm of the Jacobian than any of the normalised models but this is not the case for deeper models.
For models larger than three layers the top-performing normalised experiments exhibit a smoothness similar to that of the baseline but also increased performance.
A detailed description of the setup and results can be found in appendix~\ref{sub:app-minatar--correlation}.
In conclusion, smoothness does not fully account for the higher performance brought by \ac{sn}.

To further explore a possible connection between smoothness and performance we did several experiments with other regularisation methods. First we perform a careful tuning of \acf{gp} \cite{gulrajani2017Improving} as an alternative  method of imposing smoothness constraints on the learned action-value functions and we found it generally hurts performance and at best it makes no difference. We then considered \acf{bn} \citep{ioffe2015batch} for its ability to control the Lipschitz constant \emph{of the loss function w.r.t. the parameters} \citep{santurkar2018does}, but also generally thought to improve the optimisation properties of neural networks \citep{arora2018theoretical}. We once again found it degrades performance, which is also consistent with the \ac{drl} literature in the small batch-size regime \citep{bhatt2019crossnorm}. We give the full details in supplementary~\ref{sub:app-minatar--regularisation}. 

\subsection{Limitations of 1-Lipschitz networks and practical considerations}
\label{sub:analysis--smoothness_too_much}
Fig. \ref{fig:minatar_SN_depth_vs_width_TWO} characterises the performance of \ac{sn} as a function of the model size and the number of normalised layers. \ac{sn} generally leads to a good performance when a small number of layers are targeted.
However, normalising more than half of the number of layers in a depth-5 network results in degraded behaviour suggesting a too strong regularisation effect (see Appendix~Fig. \ref{app-fig:minatar_SN_depth_vs_width_FULL}).

We find this to be consistent with prior work showing that when ReLU networks are constrained to be 1-Lipschitz by applying \ac{sn} to all their layers, it reduces the range of functions they can express \citep{Huster2018LimitationsOT,anil2019sorting}.
Related, smoothness regularisation methods are generally employed with small coefficients in order to avoid performance penalties.
For example \acf{sr} in \citep{yoshida2017SpectralNR} uses a coefficient $\lambda = 0.01$ and the orthogonality regularisation in Parseval Networks \citep{Ciss2017ParsevalNI} uses a $\beta \in \{0.0001, 0.0003\}$ which can make the network to be far from 1-Lipschitz as discussed in \citep{anil2019sorting}.
When \ac{sn} is deployed in the adversarial robustness task \citep{gouk2020regularisation}, the optimal $\lambda_i$ values for different layer types were found in the range $\{2, .., 50\}$, a considerable departure from the 1-Lipschitz bound for the entire network. We find these reports of relaxing the 1-Lipschitz constraint consistent with our selective \ac{sn} design choice.


\subsection{Spectral Normalisation has primarily an optimisation effect}
\label{sec:analysis--optimisation}

We claim that \ac{sn} improves the performance of the value-function estimator by affecting the optimisation dynamics. Precisely, \ac{sn} provides a scheduler that adapts the size and the direction of the optimisation step as the weights increase during training. In the case of Adam \cite{kingma2014adam} \ac{sn} could be compensating for how curvature is being approximated in the late training regime by the expected squared gradient which should vanish at convergence. 

We start by comparing the gradients of the normalised neural network with those of its standard counterpart. We then build two incremental approximations to \ac{sn} which directly affect the optimisation step using the spectral radius. We provide empirical evidence that these data-dependent alterations of the optimiser recover and sometimes surpass the performance boost from \ac{sn}, supporting our optimisation interpretation of the method.

Consider a feed-forward estimator with $L$ parametric layers (for brevity we will assume linear projections, although everything applies to convolutions as well). All except the last layer are followed by rectifiers.
\begin{align}
    \act{0} &\triangleq \data \\
    \label{eq:mlp_preactivations}
    \preact{i} &= \weight{i} \act{i-1} + \bias{i} & 1 \le i \le L \\
    \act{i} &= \relu\left( \preact{i} \right) & 1 \le i < L
\end{align}
We impose that a subset of layers $\mathcal{S} \subseteq \left\lbrace 1, 2, \ldots, L\right\rbrace$ are individually $k$-Lipschitz continuous by using normalised parameters $\forall i \in \mathcal{S}: \weight[sn]{i} = \rho_{i}^{-1}\weight{i}$ with $\rho_i = \max\left( \langle\rho\left(\weight{i}\right)\rangle, k\right)$. We use $\rho\left(\cdot\right)$ to denote the spectral radius (i.e. the largest singular value of $\weight{i}$) and $\langle \cdot \rangle$ for the \emph{stop gradient} function, using hat notations for all the quantities in the normalised network: $\preact[sn]{i} = \weight[sn]{i} \act[sn]{i-1} + \bias{i}$. Since the bias is not scaled along with the weights, the sign of the pre-activations is not preserved ($\left[\preact{i} >0 \right] \neq \left[\preact[sn]{i} > 0\right]$) therefore one cannot write an explicit relation between $\dLdW[]{i}$ and $\dLdW[sn]{i}$. We will now make a slight modification to \ac{sn} such that writing such a relation becomes possible.

\begin{figure}
    \centering
    \includegraphics[width=\columnwidth]{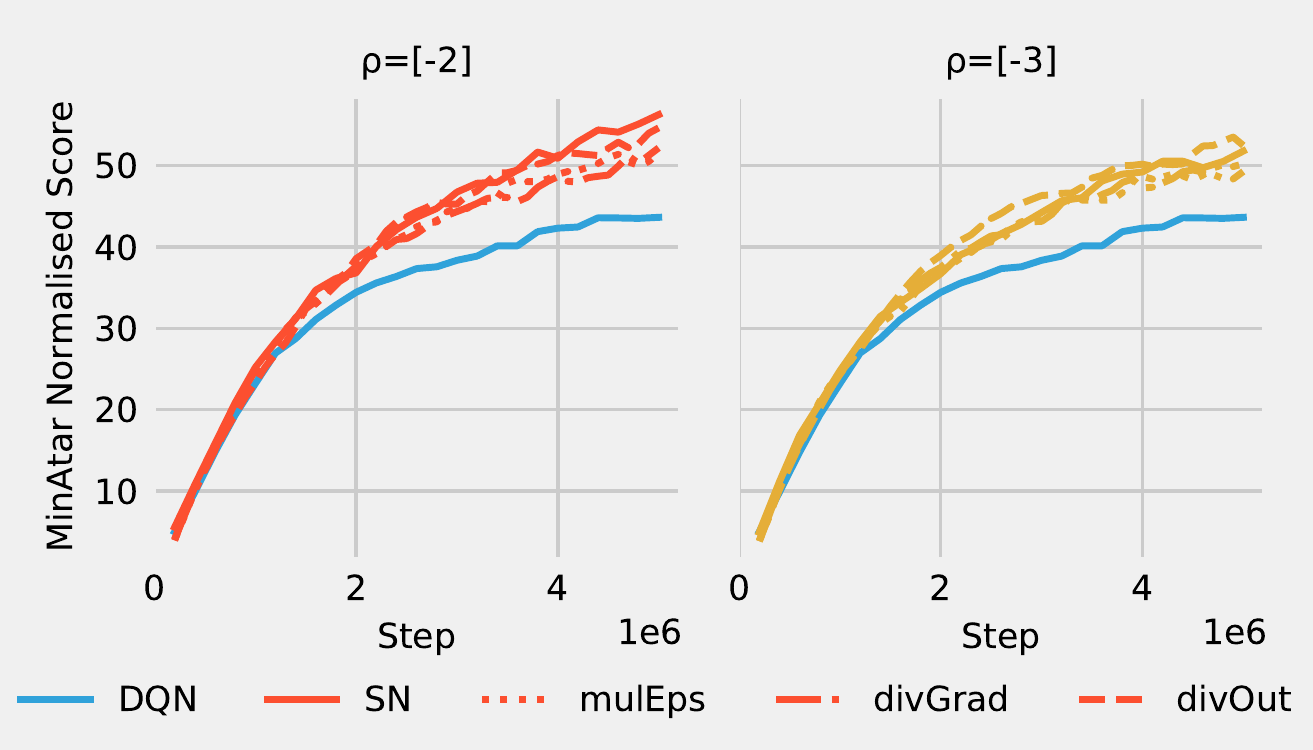}
    \caption[Schedulers]{
        \textbf{Spectral schedulers recover \ac{sn} performance.}
        Average normalised scores over MinAtar games and four different models. Refer to Fig.\ref{app-fig:schedulers_detailed} for completion.}
    \label{fig:schedulers}
\end{figure}


\paragraph{Bias scaling.} Consider the following alteration of the spectral normalisation procedure presented above. As before, we perform spectral normalisation on a subset of layers $\mathcal{S}$. In addition, for each normalised layer $l \in \mathcal{S}$ we also scale the bias terms of subsequent layers $i>l$ with $\rho_{l}^{-1}$ as shown in Eq.~\ref{eq:snpreact}. By doing so we can relate the pre-activations $\preact[sn]{i}$ on all layers to those from the unnormalised \ac{mlp} described in Eq.~\ref{eq:mlp_preactivations}:
\begin{align}
    \rho^{-1}_{i:j} &\triangleq \prod_{i \le k \le j \wedge k \in \mathcal{S}} \rho^{-1}_{k} \\
    \preact[sn]{i} &= \rho_{i}^{-1} \left(\weight{i} \act[sn]{i-1} + \rho^{-1}_{1:i-1}\bias{i} \right) \\ &= \rho_{i}^{-1}\weight{i} \act[sn]{i-1} + \rho_{1:i} \bias{i} = \rho^{-1}_{1:i} \preact{i} \label{eq:snpreact}\\
    \loss[sn] & \triangleq \lossfn\left(\preact[sn]{L}\right) = \lossfn\left(\rho^{-1}_{1:i} \preact{i}\right)
\end{align}
We can now compute the gradients w.r.t. the unnormalised parameters for both the standard estimator and the normalised one with scaled biases:
\begin{align}
    &\text{\ac{mlp}} &&\text{\ac{sn}+bias scaling} \nonumber\\
    \label{eq:w_comp}
    \dLdW{i} &= \mathbf{J}_{i} \dLdpre{L} \trn{\act{i-1}}  & \dLdW[sn]{i} &= \rho^{-1} \mathbf{J}_{i} \dLdpre[sn]{L} \trn{\act{i-1}}\\
    \label{eq:b_comp}
    \dLdb{i} &= \mathbf{J}_{i} \dLdpre{L} &  \dLdb[sn]{i} &= \rho^{-1} \mathbf{J}_{i} \dLdpre[sn]{L}
\end{align}
where $\rho^{-1} \triangleq \prod_{i \in \mathcal{S}} \rho_{i}^{-1}$; the jacobian w.r.t. estimator's outputs: $\dLdpre[sn]{L} \triangleq \dLd[sn]{\preact[sn]{L}}$; and $\mathbf{J}_{i} \triangleq \prod_{j=i}^{L-1} \left[ \diag\left(\iverson{\preact{j} > 0}\right) \trn{\weight{j+1}} \right]$.

\paragraph{Output scaling.} The right-hand expressions in Eq.~\ref{eq:w_comp}~and~\ref{eq:b_comp} show that \textsc{SN+bias scaling} is computationally equivalent with simply scaling the un-normalised network's output with $\rho^{-1}$. Doing so yields the same gradients as \textsc{\ac{sn}+bias scaling}, reducing the importance of the depths where weights are conditioned.
We refer to this approximations as \textsc{divOut}. We argue that outside some pathological cases, the bias terms easily adapt to the layer's mean activation in both cases therefore \ac{sn} and \textsc{divOut} should have similar behaviours. Indeed, experiments performed on MinAtar confirm that the \textsc{divOut} approximation recovers the performance of SN (see Figure~\ref{fig:schedulers}). Inspecting the performance on multiple architectures and normalisations we also observe that \textsc{divOut} fails to learn when \ac{sn} fails (Fig.~\ref{app-fig:schedulers_detailed} in Appendix). We therefore consider \textsc{\ac{sn} + bias scaling} (and its equivalent formulation \textsc{divOut}) a reasonable proxy to study the effects of \ac{sn}.

\paragraph{The gradient scaling effect.} Eq.~\ref{eq:w_comp},~and~\ref{eq:b_comp} show that the effect \textsc{\ac{sn}+bias scaling} has on the updates is two-fold. First, gradients are scaled by $\rho^{-1}$. Second, the model changes, therefore the jacobians of the loss function with respect to the network's output changes: $\dLdpre{L} \neq \dLdpre[sn]{L}$. We argue that specifically for DQN where the TD-error is put in a Huber cost, there is no evident scaling relation between $\dLdpre{L}$ and $\dLdpre[sn]{L}$ (for values larger than 1, the gradient is just the sign of $y-t$). We therefore isolate the first effect, i.e. scaling the gradients with the inverse spectral radius before passing it to the gradient based optimiser, in a scheduler we name \textsc{divGrad}.

We again evaluate this scheduler using multiple architectures and sets of spectral radii on the games in MinAtar. The averaged learning curves in Figure~\ref{fig:schedulers} demonstrate that this gradient scaling effect of {\it \ac{sn} is sufficient to explain the performance gain, indicating that \ac{sn} actually proposes better optimisation steps rather than conditioning the network in a better subspace}.

\begin{figure}
    \centering
    \includegraphics[width=\columnwidth]{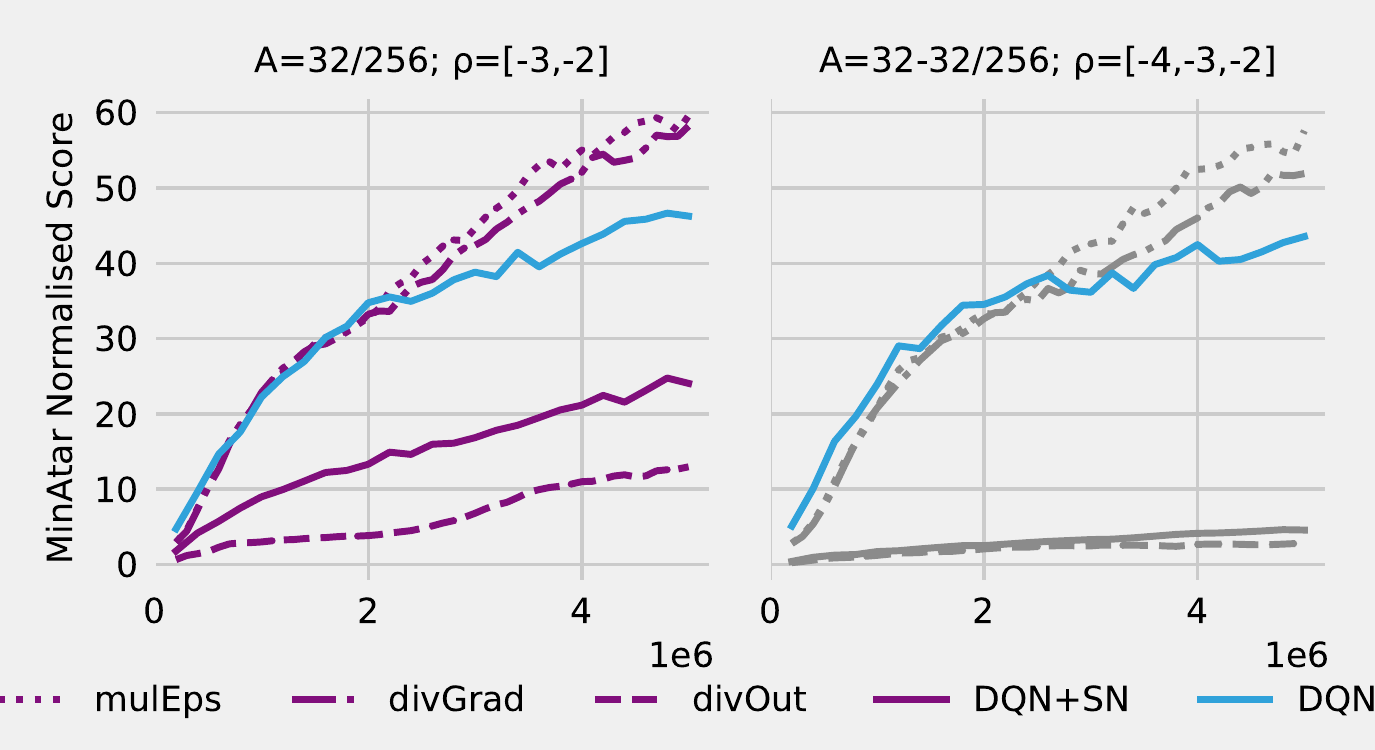}
    \caption[Schedulers]{\textbf{
        Spectral schedulers improve performance even when the radii of all layers are used.}
        In these cases \ac{sn} fails to train. 
        Minatar Normalised Score for two models.
        Refer to Fig.\ref{app-fig:schedulers_detailed} for completion.}
    \label{fig:all_radii_schedulers}
\end{figure}

We also notice that \textsc{divGrad} improves the performance of DQN event when the spectral norms of all hidden layers are considered, case in which \ac{sn} fails to train (Figure~\ref{fig:all_radii_schedulers}). 

\paragraph{Interactions with Adam.}

In order to account for the surprising performance of \textsc{divGrad}, we study its effect when the optimiser used is Adam \cite{kingma2014adam} (Eqs.~\ref{eq:adam_momentum},~\ref{eq:adam_snd_order},~\ref{eq:adam_update}), the recent\footnote{DQN and its immediate successors used RMSprop.} first choice for optimising \ac{drl} algorithms. To do that we compare an optimisation step in a regular \ac{mlp} with the same quantity when gradients are inversely scaled with the spectral radius.
\begin{align}
    \mathbf{v}_{t+1} &\longleftarrow \beta_1 \mathbf{v}_{t} + \left(1 - \beta_1\right) \mathbf{g}_{t} \label{eq:adam_momentum} \\
    \mathbf{s}_{t+1} &\longleftarrow \beta_2 \mathbf{s}_{t} + \left(1 - \beta_2\right) \mathbf{g}^{2}_{t} \label{eq:adam_snd_order} \\
    \Delta\weight{t} &\triangleq \eta \frac{\left(1 - \beta_1^t\right)^{-1}\mathbf{v}_{t+1}}{\sqrt{\left(1 - \beta_2^t\right)^{-1}\mathbf{s}_{t+1}}+ \epsilon}\label{eq:adam_update}
\end{align}
The spectral radius changes slowly during training (Fig.\ref{app-fig:dqn-minatar-all-radii-2} in Appendix), making it reasonable to approximate the exponentially decayed sums in $\hat{\mathbf{v}}_{t}$, and $\hat{\mathbf{v}}_{t}$:
\begin{align}
    \mathbf{v}^{'}_{t+1} &\approx \rho_{t}^{-1} \mathbf{v}_{t+1} \label{eq:adam_scaled_momentum} \\
    \mathbf{s}^{'}_{t+1} &\approx \rho_{t}^{-2} \mathbf{s}_{t+1} \label{eq:adam:scaled_snd_order}\\
    \Delta\weight{t}^{'} &\approx \eta \frac{\left(1 - \beta_1^t\right)^{-1}\mathbf{v}_{t+1}}{\sqrt{\left(1 - \beta_2^t\right)^{-1}\mathbf{s}_{t+1}}+ \rho_t \epsilon}\label{eq:eps_update}
\end{align}
It means that (in a instantaneous comparison with the gradient for the unregularized network) the step size when applying gradient scaling with a slowly increasing $\rho$ reduces the learning rate while making the projection closer and closer to SGD with momentum. Or, equivalently, as learning progresses the curvature approximation in the denominator is dominated by the uniform scale factor $\rho\epsilon$. We therefore propose a scheduler (\textsc{mulEps}) where we multiply the $\epsilon$ term by the product of spectral radii as in Eq.~\ref{eq:eps_update}
, expecting a close behaviour to that of \textsc{divGrad}. Experiments on MinAtar confirm this, providing a possible interpretation for the efficiency of \ac{sn}, namely that it modulates $\epsilon$. 

While in \ac{sl} $\epsilon$ is mostly used for numerical stability, its role in dealing with unreliable approximation of the curvature has been considered.
Since gradient statistics are typically assumed to be unreliable early in training~\citep{liu2019regularization}, a decreasing schedule for the damping factor was suggested in \citep{martens2015optimizing}. 
However these heuristics do not seem to match the needs of \ac{drl}. 
This is emphasised by the typical use of much higher $\epsilon$ in \ac{drl}~\citep{bellemare2017distributional,mnih2015human} and by our findings, suggesting the modulation of $\epsilon$.


\section{Related work}
\label{sec:related_work}

\ac{sn} and its penalty term formulation \acf{sr} \cite{yoshida2017SpectralNR}, have been mostly researched in the context of regularisation for example as a defense against adversarial examples \citep{tsutsuku2018,farnia2018generalizable,gouk2020regularisation}, to improve the stability of \ac{gan} training \citep{kurach2019large} or to improve robustness of uncertainty estimates \citep{liu2020simple}.
It was recently used in \ac{drl}~\citep{yu2020mopo} also in the context of robust uncertainty estimation for model based RL. 
However the optimisation effect investigated in Sec.~\ref{sec:analysis} has not been considered. 
\cite{rosca2020case} provides a thorough overview of the current understanding of smoothness constraints, applications and methods, including \ac{sn}.

Probably the most related method that has been extensively studied from an optimisation perspective is \ac{wn} \citep{salimans2016weight}.
Similary to \ac{sn} they rely on a hard projection
$\weight[sn]{i} = g \mathbf{v} / \lVert \mathbf{v} \rVert_F$,
with the exception that $g$ is learned and the use of a simpler Frobenius matrix norm. 
The authors 
also report results on a subset of \ac{ale} environments for DQN, showing gains over the baseline.
%
\cite{miyato2018spectral} contrasts \ac{sn} and \ac{wn} and argue that the Frobenius norm encourages a loss in the number of usable features of the learned representations. 
Our experiments support this argument: measuring the \emph{effective rank} \citep{kumar2020implicit} shows a faster loss of feature rank for the baseline agent compared to any \ac{sn} agent (Fig.~\ref{app-fig:rank-atari} in Appendix).
Since \ac{wn} fuels a loss in rank we believe this might explain the performance difference between the two.
%

\ac{sn} together with \ac{wn} and \acf{bn}~\cite{ioffe2015batch}, are part of a larger family of normalisation methods that make the layer invariant to change in the scale of the parameters.
In \citep{van2017l2,hoffer2018norm,arora2018theoretical} the authors study the effect of L2 regularisation in \ac{bn} and \ac{wn} networks deriving an \emph{effective learning rate} $\eta' = \eta / \lVert \weight{} \rVert$ that depends on the evolution of the weight norm.
They also show that the norm interacts with the damping factor in adaptive algorithms, including Adam, such that $\epsilon' = \epsilon \lVert \weight{} \rVert$ effectively behaving as a scheduler on these hyper-parameters.
Our analysis in Sec.~\ref{sec:analysis--optimisation} looks at the relation between the regular and normalised update steps and identifies a similar scaling factor of the damping term, while preserving the learning rate.

The damping term $\epsilon$ has been often overlooked in \ac{sl} and relegated only to numerical stability, however in \ac{drl} it is usually set rather high in both Adam and RMSprop without a justification.
We believe part of the gains we report is due to the implicit scheduling of $\epsilon$.
We corroborate this finding with work on accurate approximation of the curvature of the loss function where scheduling the Tikhonov damping is proposed.
For example, in K-FAC, 
the damping value starts high and it is decreased as the information about curvature improves~\citep{martens2010deep,martens2015optimizing}.
%
The $\epsilon$ term in Adam has a similar purpose but on a weaker approximation of the curvature based on the Empirical Fisher Information Matrix \citep{Pascanu+Bengio-ICLR2014,kunstner2019limitations}.
Our study in Sec.~\ref{sec:analysis--optimisation} suggests that in \ac{drl} this approximation is even more deficient and does not improve during training, as in general we see a benefit from the monotonic increase of the damping factor.

Recent works address the optimisation issues specific to TD learning.
\citep{bengio2020InterferenceAG} discusses the negative interplay between the adaptive methods from \ac{sl} (RMSprop, Adam) and TD(0).
Others propose optimisers for RL, but they are either restricted to linear estimators \citep{givchi2014QuasiNT,sun2020adaptive}, or they show no empirical advantages \citep{romoff2020TDpropDJ}.
\section{Conclusion}

We identify and characterise the strong performance boost when applying \acf{sn} on the network \ac{drl} agents use for state-action value estimation.
%
%
Through careful ablations on a large number of architectures and settings we disentangle between smoothness and optimisation effects.
We find that rather than restricting the irregularities of the network function, \ac{sn} changes the optimisation dynamics providing a good scheduler.
We also derive explicit schedulers by moving the spectral norm in the update step of Adam and show that these recover the performance of \ac{sn}.
%
%
These observations suggest there is much to gain by designing better adapting optimisers for \ac{drl}.

\cleardoublepage


\bibliography{bibliography}
\bibliographystyle{icml2021}

\cleardoublepage
\appendix

\section{Methods}
\label{sec:app-methods}

\subsection{Computing the spectral norm}
\label{sec:the-norm}

\paragraph{Power Iteration.} Computing the dominant singular value at each step would be expensive, therefore spectral normalisation is usually performed through power iteration. For each set of weights $\weight{i} \in \mathbb{R}^{N_{i} \times N_{i-1}}: i \in \mathcal{S}$ the corresponding left and right singular vectors are stored, and two extra matrix-vector multiplications for each normalised layer are performed at each forward pass (see Formulas~\ref{eq:piter_v},~\ref{eq:piter_u}). At inference time there is no extra computational cost.
\begin{align}
    \label{eq:piter_v}
    \mathbf{v} &\leftarrow \weight{i}\mathbf{u}^{(t-1)}; & \alpha &\leftarrow \lVert \mathbf{v}\rVert; & \mathbf{v}^{(t)} &\leftarrow \alpha^{-1} \mathbf{v} \\
    \label{eq:piter_u}
    \mathbf{u} &\leftarrow \trn{\weight{i}}\mathbf{v}^{(t)}; & \rho &\leftarrow \lVert \mathbf{u}\rVert; & \mathbf{u}^{(t)} &\leftarrow \rho^{-1} \mathbf{u}
\end{align}
For convolutional layers we adapt the procedure in \cite{gouk2020regularisation} and the two matrix-vector multiplications are replaced by convolutional and transposed convolutional operations.
\paragraph{Backpropagating through the norm.} Since the parameters that are tuned during optimisation are the unnormalised weights $\weight{i}$, we investigated if there are any advantages in backpropagating through the power iteration step, i.e. considering the partial derivative $\frac{\partial \weight[sn]{i}}{\partial \rho_{i}} \frac{\partial \rho_i}{\partial \weight{i}}$ when computing the gradient. Precisely, we verified if dropping the second term in the right-hand side of Formula~\ref{eq:norm_jac} biases the gradient.
\begin{align}
    \label{eq:norm_jac}
    \frac{\partial \loss}{\partial \weight{i}} = \rho_{i}^{-1}\frac{\partial \loss}{\partial \weight[sn]{i}} - \rho_{i}^{-2} \mathbf{u}\trn{\mathbf{v}}\left(vec\left(\frac{\partial \loss}{\partial \weight[sn]{i}}\right)vec\left(\weight{i}\right)\right)
\end{align}
In a batch of experiments performed on a subset of games of Atari we noticed no loss in performance when dropping the Jacobian from the power iteration.

\subsection{Computing the norm of the Jacobians}

Experiments in Sec.~\ref{sub:analysis--smoothness} used the maximum norm of the jacobians w.r.t. the inputs as an indirect metric for network function's smoothness. For each network we collected thousands of states (\emph{on-policy}), computing the maximum euclidean norm of the jacobian w.r.t. the inputs: $\max_{i, \mathbf{x}}\vert\vert \frac{\partial q_i\left(\mathbf{x};\theta\right)}{\partial \mathbf{x}}\vert\vert_2$. Note that we used the euclidean norm (and not the operator norm) to measure smoothness.

\clearpage
\section{MinAtar experiments}
\label{sec:app-minatar}

\paragraph{Game selection.} MinAtar \cite{young19minatar} benchmark is a collection of five games that reproduce the dynamics of \acf{ale} counterparts, albeit in a smaller observational space. Out of \emph{Asterix, Breakout, Seaquest, Space Invaders} and \emph{Freeway} we excluded the latter from all our experiments since all the agents performed essentially the same on this game.

\paragraph{Network architecture.} All experiments on MinAtar are using a convolutional network with $L_C$ convolutional layers with the same number of channels, a hidden linear layer, and the output layer. The number of input channels is game-dependent in MinAtar. All convolutional layers have a kernel size of $3$ and a stride of $1$. All hidden layers have rectified linear units. Whenever we vary depth we change the number of convolutional layers $L_C$, keeping the two linear layers. When the width is varied we change the width of both convolutional layers (e.g. 16/24/32 channels) and the penultimate linear layer. All convolutional layers are always identically scaled.

We list all the architectures used in various experiments described in this section in Table~\ref{tbl:app-ALL-models}.

 \paragraph{General hyper-parameter settings.} In all our MinAtar experiments we used the same set of hyper-parameters returned by a small grid search around the initial values published by \cite{young19minatar}. We list the values we settled on in Table~\ref{tbl:app-minatar_hyperparameters}. For the rest of this section we only mention how we deviate from this set of hyper-parameters and settings for each of the experiments that follow.

\begin{table}[h]
    \vskip 0.15in
    \begin{center}
        \begin{small}
            \begin{tabular}{lcc}
                \toprule
                \textsc{Hyper-parameter}        & \textsc{Value} \\
                \midrule
                \rule{0pt}{2ex}%
                discount $\gamma$               & \textsc{0.99} \\
                update frequency                & \textsc{4} \\
                target update frequency         & \textsc{4,000} \\
                \midrule
                \rule{0pt}{1ex}%
                starting $\epsilon$             & \textsc{1.0} \\
                final $\epsilon$                & \textsc{0.01} \\
                $\epsilon$ steps                & \textsc{250,000} \\
                $\epsilon$ schedule             & linear \\
                warmup steps                    & \textsc{5,000} \\
                \midrule
                \rule{0pt}{1ex}%
                replay size                     & \textsc{100,000} \\
                history length                  & \textsc{1} \\
                \midrule
                \rule{0pt}{1ex}%
                cost function                   & \textsc{MSE Loss} \\
                optimiser                       & \textsc{Adam} \\
                learning rate $\eta$            & \textsc{0.00025} \\
                damping term $\epsilon$         & \textsc{0.0003125} \\
                $\beta_1, \beta_2$              & \textsc{(0.9, 0.999)} \\
                \midrule
                \rule{0pt}{1ex}%
                validation steps                & \textsc{125000} \\
                validation $\epsilon$           & \textsc{0.001} \\
                \bottomrule
            \end{tabular}
        \end{small}
    \end{center}
    \caption{
        MinAtar general hyper-parameter settings.
    }
    \label{tbl:app-minatar_hyperparameters}
\end{table}

\paragraph{MinAtar Normalised Score.} In our work we present many MinAtar experiments as averages over the four games we tested on. Since in MinAtar the range of the expected returns is game dependent we normalise the score. Inspired by the Human Normalised Score in \cite{mnih2015human} we take the largest score ever recorded by a baseline agent in our experiments and use it to compute $\text{MNS} = 100 \times (\text{score}_\text{agent} - \text{score}_\text{random}) / (\text{score}_\text{max} - \text{score}_\text{random})$. We can then use the resulting MinAtar Normalised Score whenever we need to report performance aggregates over the games.

\begin{table}[h]
    \vskip 0.15in
    \begin{center}
        \begin{small}
            \begin{tabular}{lcc}
                \toprule
                \textsc{Game}   & \textsc{Max}    & \textsc{Random} \\
                \midrule
                \rule{0pt}{2ex}%
                Asterix         & \textsc{78.90}     & \textsc{0.49} \\
                Breakout        & \textsc{122.88}    & \textsc{0.52} \\
                Seaquest        & \textsc{93.91}     & \textsc{0.09} \\
                Space Invaders  & \textsc{360.92}    & \textsc{2.86} \\
                \bottomrule
            \end{tabular}
        \end{small}
    \end{center}
    \caption{
        MinAtar maximum and random scores used for computing the MinAtar Normalised Score.
    }
    \label{tbl:app-minatar_normalization_scores}
\end{table}

\subsection{Large optimiser hyper-parameter sweep}
\label{sub:app-minatar--optimisation_sweep}

For the large optimiser hyper-parameter sweep we train a DQN agent with six different architectures as listed in Table~\ref{tbl:app-ALL-models}.
For each architecture we then trained on all combinations of optimisation hyper-parameters (from the lists below), both normalised and baseline agents, two seeds each.
We generated values for the learning rate using a geometric progression in the following ranges:

\begin{table}[h]
    \vskip 0.15in
    \begin{center}
        \begin{small}
            \begin{tabular}{ll}
                \toprule
                Optimiser           & Hyper-parameters \\
                \midrule
                \rule{0pt}{2ex}%
                \textsc{Adam}       & \makecell[l]{
                    $\eta$: $\{0.00001, ..., 0.00215\}$ \\
                    $\epsilon$: $\{0.00001, ..., 0.01\}$} \\
                \midrule
                \rule{0pt}{2ex}%
                \textsc{RMSprop}    & \makecell[l]{
                    $\eta$: $\{0.00001, ..., 0.001\}$ \\
                    $\epsilon$: $\{0.00001, ..., 0.0316\}$ \\
                    $\alpha \in \{0.95\}$ \\
                    centered: yes%
                } \\
                \midrule
                \rule{0pt}{2ex}%
                \textsc{Adam}       & \makecell[l]{
                    $\eta$: $\{0.00001, ..., 0.001\}$
                }\\
                \bottomrule
            \end{tabular}
        \end{small}
    \end{center}
    \caption{
        Hyper-parameters ranges for the large optimiser sweep.
    }
    \label{tbl:app-optim_sweep_hyperparams}
\end{table}

\begin{figure*}[ht]
    \centering
    \includegraphics[width=\textwidth]{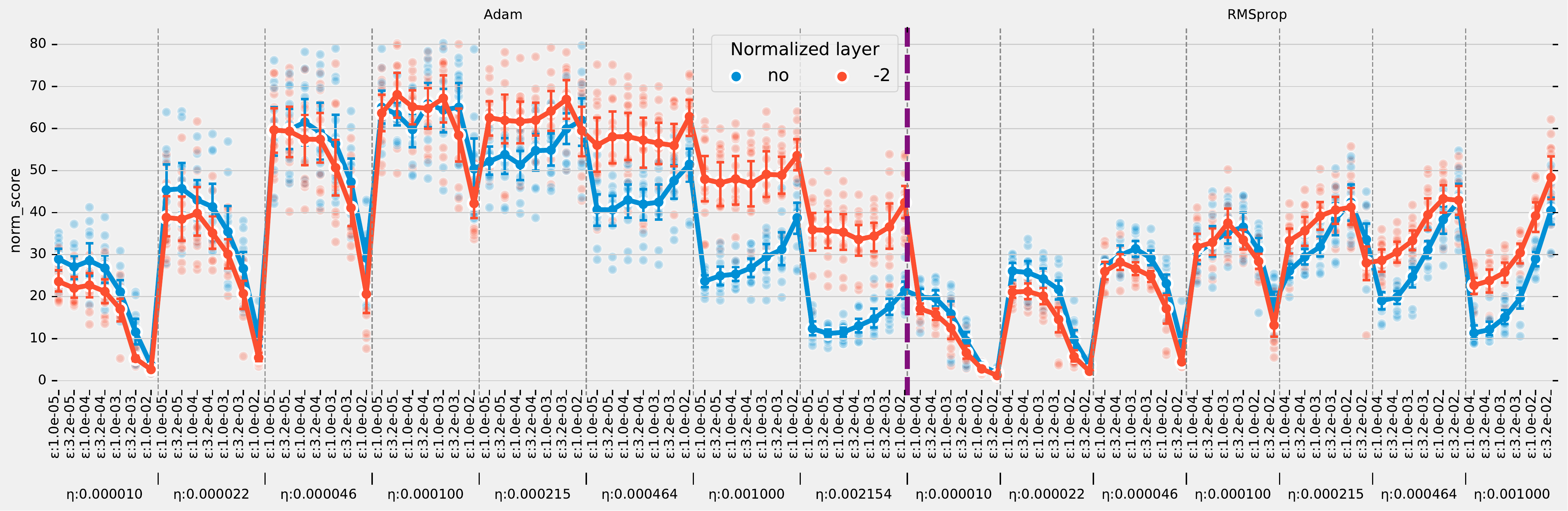}
    \caption{
        \textbf{Spectral Normalisation increases the range of effective optimisation settings.}
        Each configuration is used to train 6 models of different depth and width on four MinAtar games, two seeds each.
        Dots represent the average maximum normalised
        score over the four games.}
    \label{app-fig:minatar_optim_sorted_optim}
\end{figure*}

\begin{figure*}[ht]
    \centering
    \includegraphics[width=\textwidth]{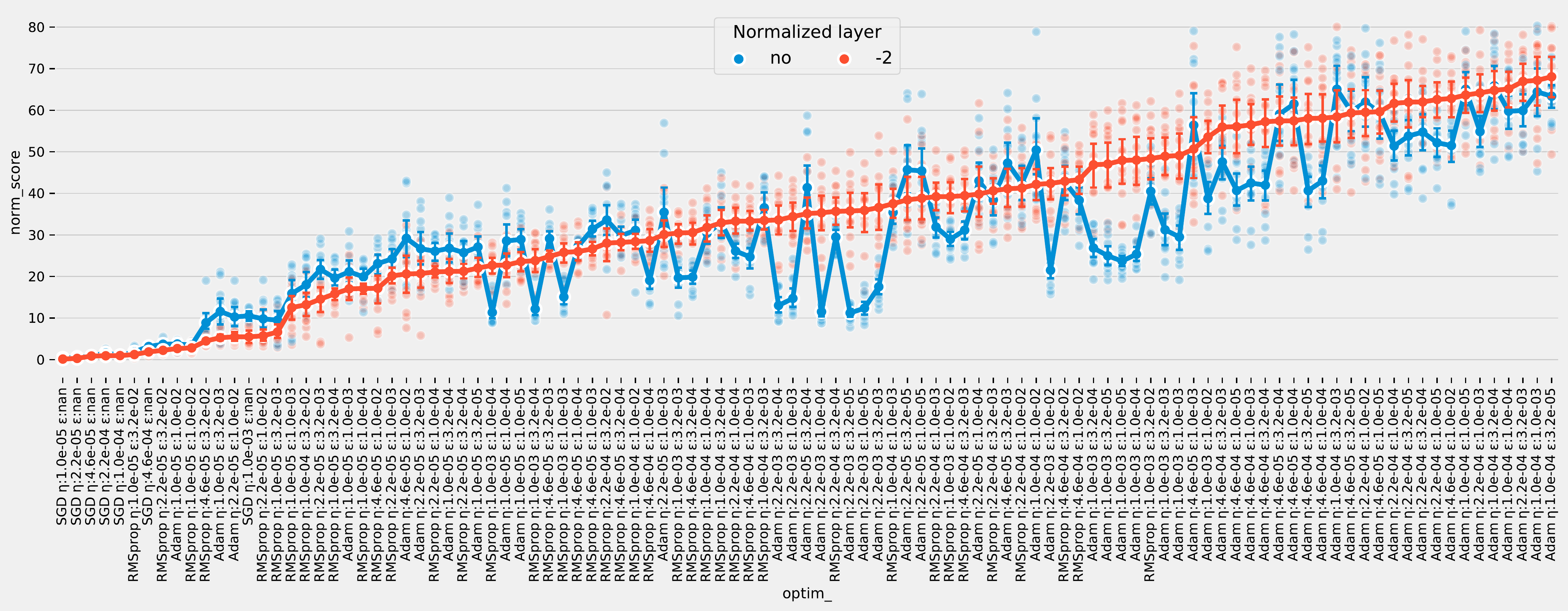}
    \caption{
        \textbf{Spectral Normalisation improves on the baseline on a wide range of optimisation settings.}
        Each optimisation setting is used to train 6 models of different depths and widths on
        four MinAtar games, two seeds each.
        Dots represent the average MinAtar Normalised Score achieved over four games.
        We sort by the mean performance of the \ac{sn}
        experiment and show the baseline stays mostly under this curve in the region of high performance optimiser configurations.}
    \label{app-fig:minatar_optim_sorted_sn}
\end{figure*}

Figure~\ref{app-fig:minatar_optim_sorted_optim} illustrates our findings. In the case of Adam optimiser we can see that normalising one layer does not degrade the performance of the baseline and allows for larger learning rates. A similar observation can be made for RMSprop and exploring learning rates larges than $0.001$ should clarify to what degree the trend already visibile continues.

Figure \ref{app-fig:minatar_optim_sorted_sn} is a different visualisation of the same results. Here we sort the x-axis by the mean normalised score of an optimiser configuration for the DQN agents equipped with \ac{sn} and plot both the performance of the baseline and the normalised experiments. We show that the performance of the normalised agent generally outperforms the  baseline, especially in the right half of the plot corresponding with higher performance agents.


\subsection{Effect on model capacity}
\label{sub:app-minatar--model_capacity}
\setcounter{figure}{13} 
\begin{figure}[ht]
    \centering
    \includegraphics[width=\columnwidth]{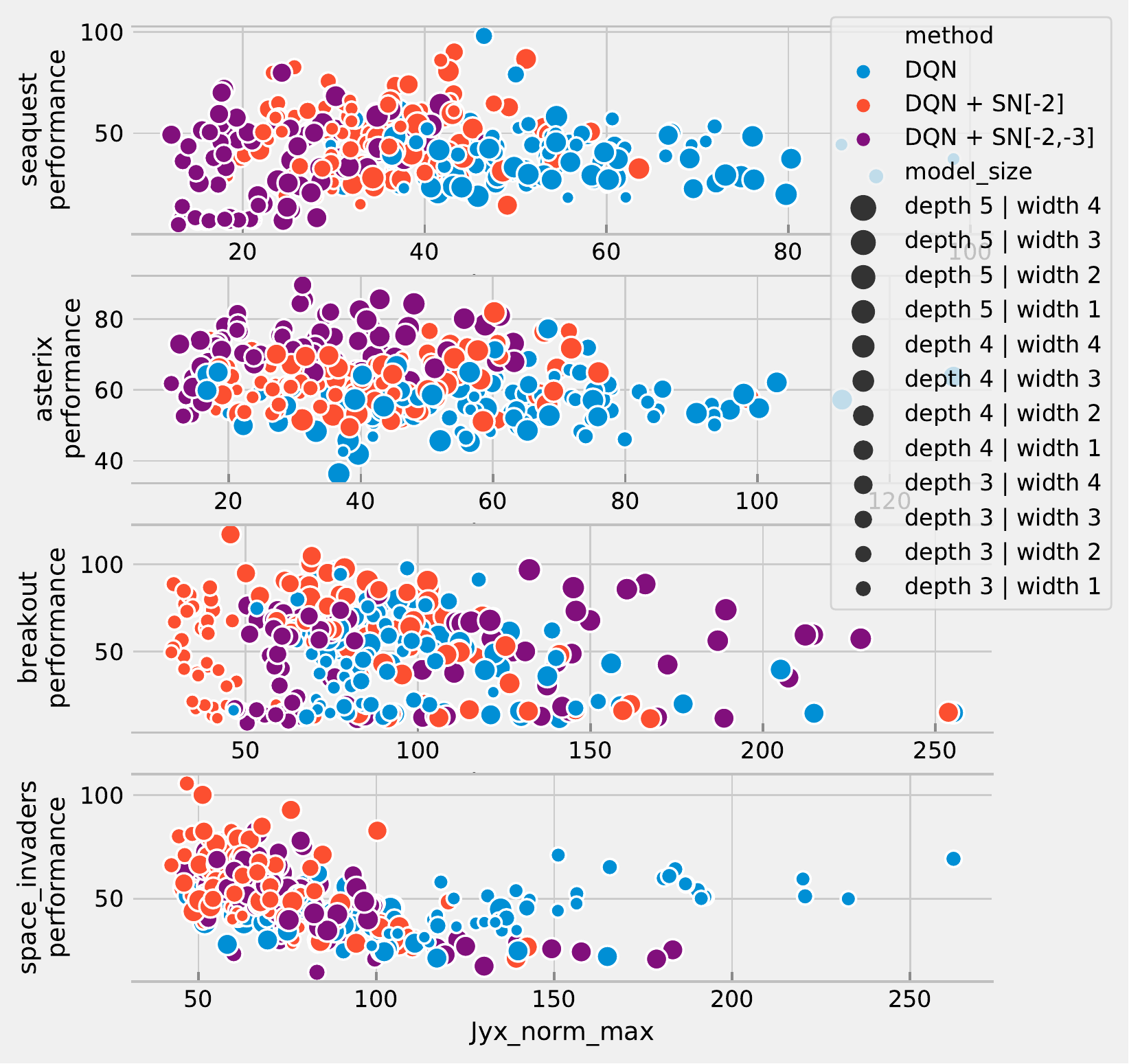}
    \caption[Baseline vs Spectral Norm]{\textbf{Applying \ac{sn} on a layer subset does not consistently produce smoother networks.} Often normalising a subset of the network's layers makes the network less smooth than the baseline while performance improves still. Each point in the graph represents the maximum performance achieved by a single seed. Detailed view of Fig.~\ref{fig:minatar_smoothness_vs_perf_vs_size}}.
    \label{app-fig:minatar_smoothness_vs_perf_vs_size_FULL}
\end{figure}

For understanding the effect of \ac{sn} on model capacity we train DQN agents with 12 different model sizes of three different depths and four different widths.
We apply \ac{sn} on various layer subsets and report in Figure~\ref{app-fig:minatar_SN_depth_vs_width_FULL} the MinAtar Normalised Score averaged over the four games.
All the other parameters remain the same as described at the beginning of this section.
Each resulting game-architecture combination was trained on $10$ seeds.


\subsection{Smoothness and performance are weakly correlated}
\label{sub:app-minatar--correlation}

In Figure~\ref{app-fig:minatar_smoothness_vs_perf_vs_size_FULL} we plot the peak performance and the norm of the Jacobian for each of the seeds in the experiment we discuss in Section~\ref{sub:app-minatar--correlation} instead of averages.

Computing a correlation measure for all the normalisation schemes in the experiment is complicated by the fact that any selection we make affects the correlation we want to measure.
Limiting ourselves to just the baseline and \textsc{DQN[-2]}, the normalisation scheme we have shown repeatedly that it does not hurt performance, we computed the Spearman rank-order correlation we report in Table~\ref{tbl:app-minatar_correlation}.

\begin{table}[h]
    \vskip 0.15in
    \begin{center}
        \begin{small}
            \begin{tabular}{lc}
                \toprule
                \textsc{Game}   & \textsc{Spearman Rank}\\
                \midrule
                \rule{0pt}{2ex}%
                Asterix         & \textsc{-0.129}\\
                Breakout        & \textsc{-0.199}\\
                Seaquest        & \textsc{-0.151}\\
                Space Invaders  & \textsc{-0.453}\\
                \bottomrule
            \end{tabular}
        \end{small}
    \end{center}
    \caption{
        Correlation between the norm of the Jacobian and peak performance for each game.
    }
    \label{tbl:app-minatar_correlation}
\end{table}

\subsection{Other regularisation methods}
\label{sub:app-minatar--regularisation}
\setcounter{figure}{15} 
\begin{figure}[ht]
    \centering
    \includegraphics[width=\columnwidth]{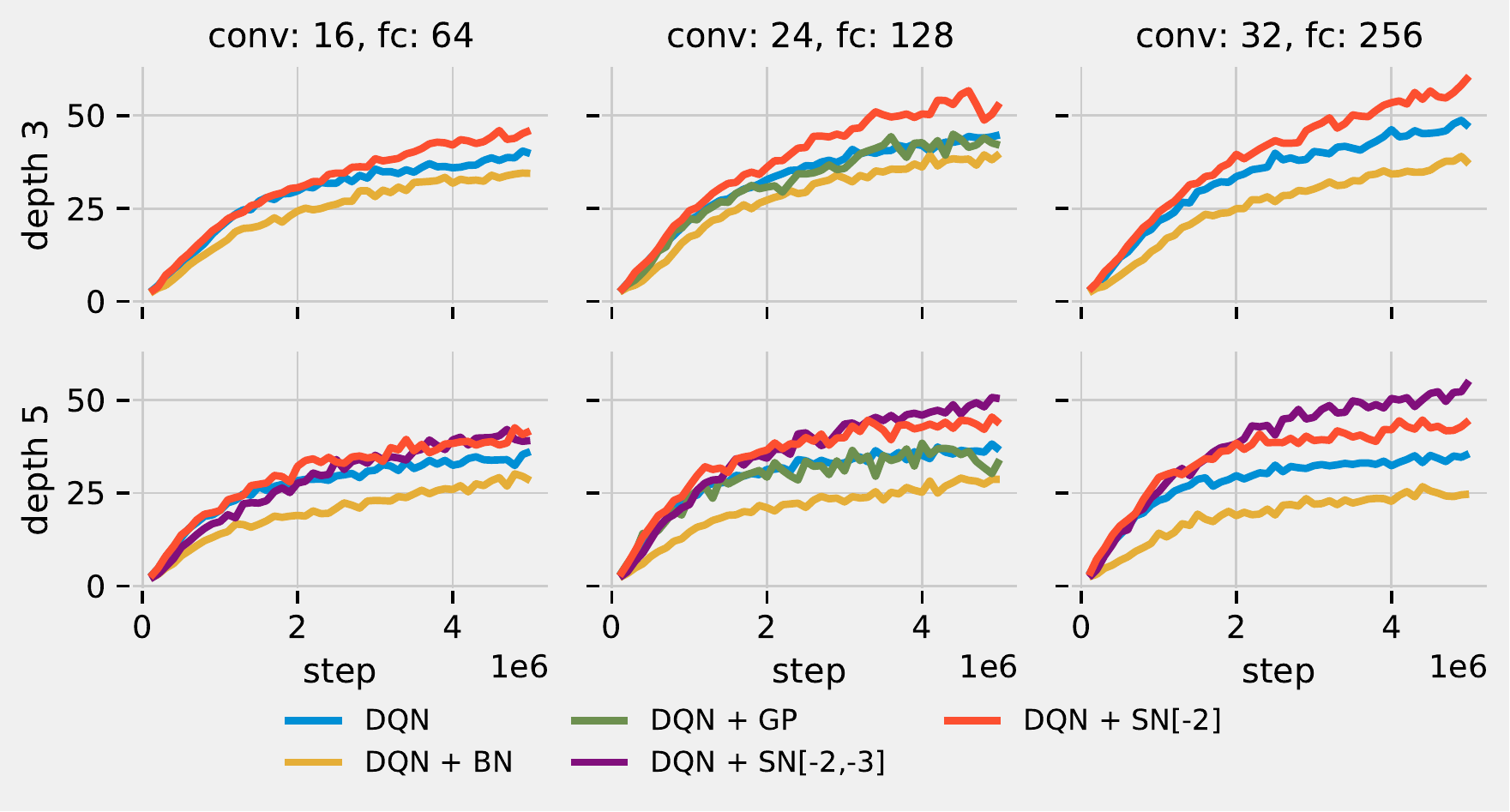}
    \caption{\textbf{Regularization does not recover Spectral Normalization performance.} Performance on MinAtar games of \ac{sn}, \ac{gp} and \ac{bn}. Each line is an average over normalized scores of each game. Ten seeds for each configuration.}
    \label{app-fig:minatar_REG_depth_vs_width}
\end{figure}

As briefly touched upon in Section~\ref{sub:analysis--smoothness}, we investigated whether other regularisation methods imposing smoothness constraints can have similar effects on the agent's performance. To this end we ran experiments with both \acf{gp} and \acf{bn} on several architectures (Table~\ref{tbl:app-ALL-models}).

\paragraph{Batch Normalisation.} For each of the architectures we employed \acf{bn} after the ReLU activation of every convolutional or linear layer except the output.

\paragraph{Gradient Penalty.} We did extensive experimentation and penalty coefficient tuning for Gradient Penalty regularisation. Specifically we tried penalising the norm of the sum of the gradients of all actions (the way \ac{gp} is usually implemented in other domains), regularising the expected norm of each Q-value with respect to the state and also regularising the norm of the gradient of the Q-value associated with the optimal action. In all cases we swept through a wide range of penalty coefficients $\lambda$ with various degree of success. In Figure~\ref{app-fig:minatar_REG_depth_vs_width} we report the results of the best setting we could identify.

\paragraph{Relaxations to 1-Lipschitz normalisation.} We run a small experiment with a three layer network to investigate the relaxation introduced by \cite{gouk2020regularisation}: $\weight[sn]{i} = \weight{i} / \max(\lambda_i, \lVert \weight{i} \rVert_2)$. %
Figure~\ref{app-fig:minatar_lipschitz_k} shows that for increased $\lambda_i$ values (which we keep equal for every layer) we are able to get good performance even when normalising all the layers of the network, further confirming our initial observation that controlling the amount of regularisation is important to achieving optimal performance and that achieving 1-Lipschitz functions is not critical in this setup. The increased computation required when approximating $\rho$ for all the layers and the addition of one hyper-parameter per layer determined us to not pursue this setup further.

\setcounter{figure}{9} 
\begin{figure}[h]
    \centering
    \includegraphics[width=\columnwidth]{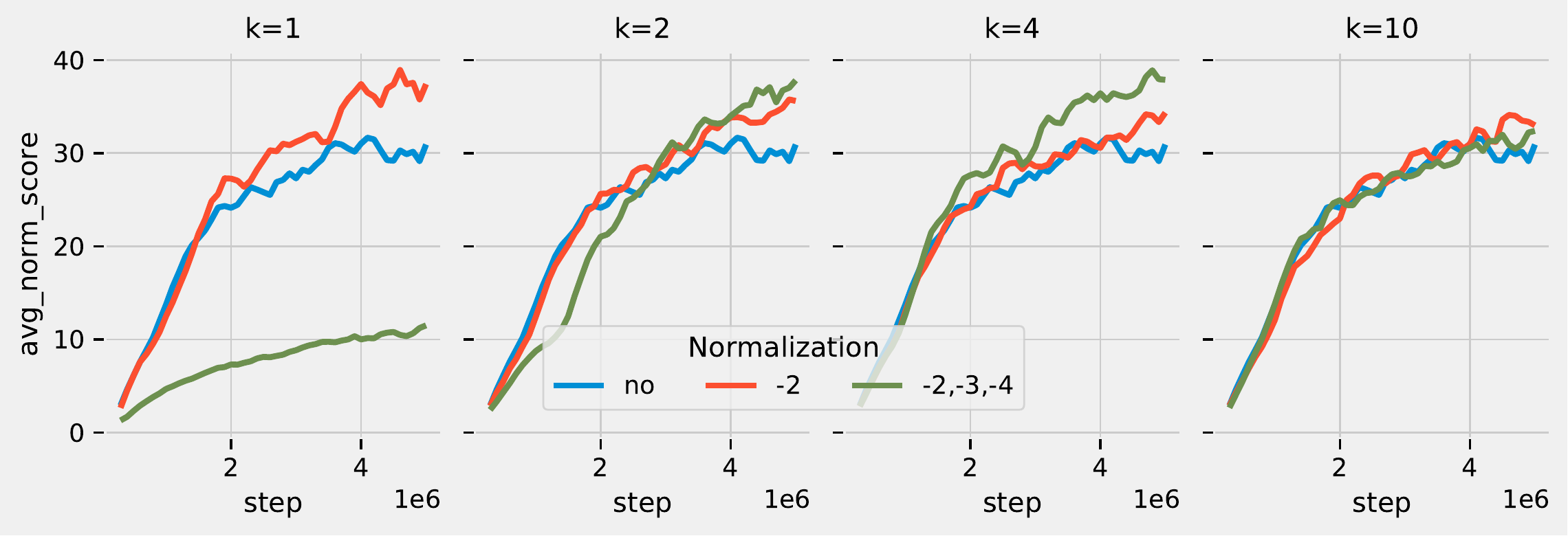}
    \caption{
        \textbf{Relaxing the 1-Lipschitz condition recovers the performance for fully normalised networks}.
        Average MinAtar Normalised Score of \ac{sn} with different target Lipschitz constants.
        Each line is an average over normalised scores of 4 games $\times$ 10 seeds.}
    \label{app-fig:minatar_lipschitz_k}
\end{figure}

\clearpage

\setcounter{figure}{11} 
\begin{figure*}[ht]
    \centering
    \includegraphics[width=\textwidth]{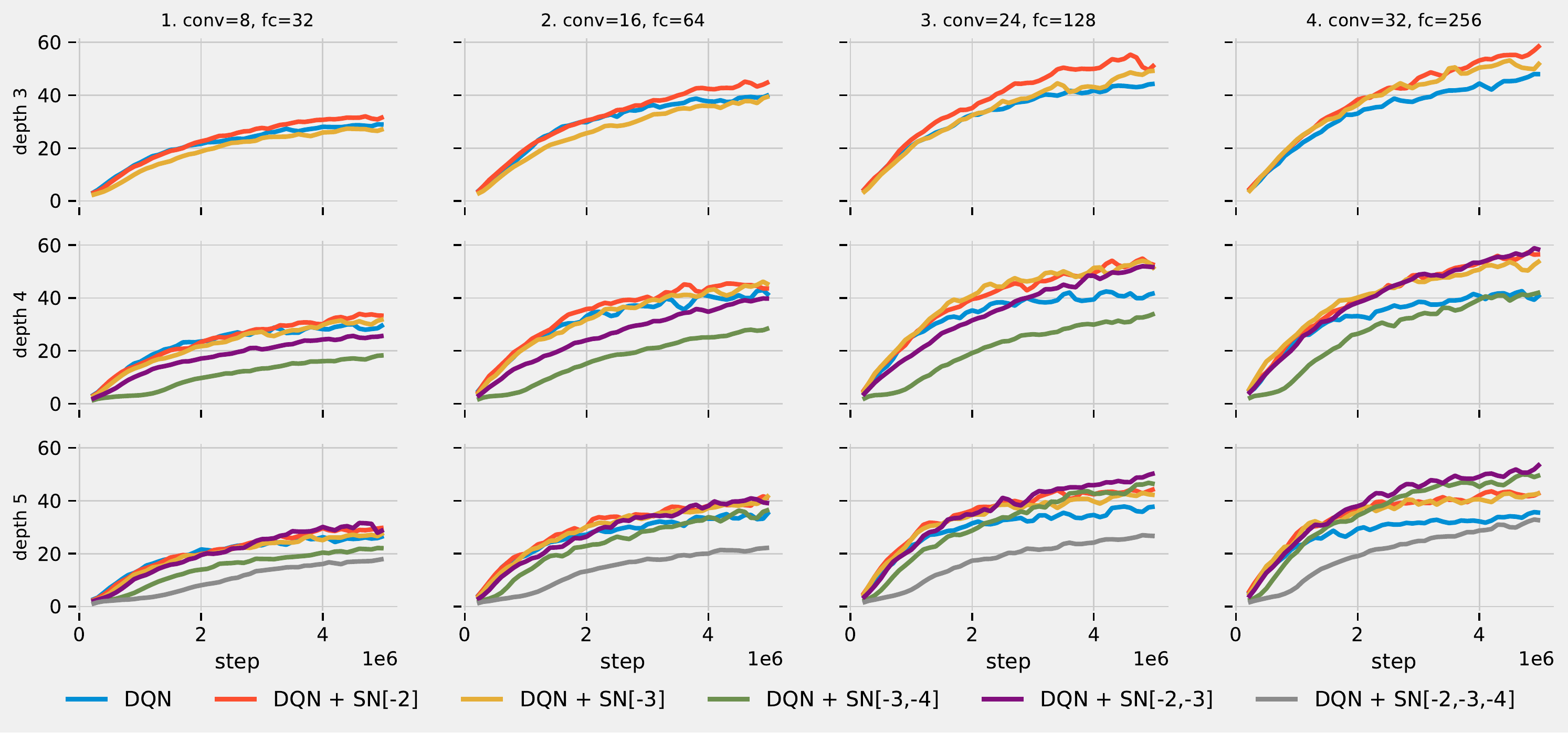}
    \caption{\textbf{Spectral Normalization shows gains for all model sizes.} Looking at the baseline (\legend{solid,line width=1.5pt,538blue} DQN), we observe two performance regimes on MinAtar: for shallow, depth 3 models, performance increases with the width of the model; for deeper models performance generally stagnates with increasing depth and width. In both regimes applying SN on individual (\legend{solid,line width=1.5pt,538red}, \legend{solid,line width=1.5pt,538yellow}) or multiple (\legend{solid,line width=1.5pt,538purple}) layers improves upon the baseline suggesting a regularisation effect we could not reproduce with other regularisation methods. Notice that the strong regularisation resulted from applying \ac{sn} to input layers (\legend{solid,line width=1.5pt,538green}) or too many layers (\legend{solid,line width=1.5pt,538grey}) can however degrade performance. Each line is an average over normalized scores of 4 games $\times$ 10 seeds. Detailed view of Fig.~\ref{fig:minatar_SN_depth_vs_width_TWO}.}
    \label{app-fig:minatar_SN_depth_vs_width_FULL}
\end{figure*}
\setcounter{figure}{10} 
\begin{figure*}[ht]
    \centering
    \includegraphics[width=\textwidth]{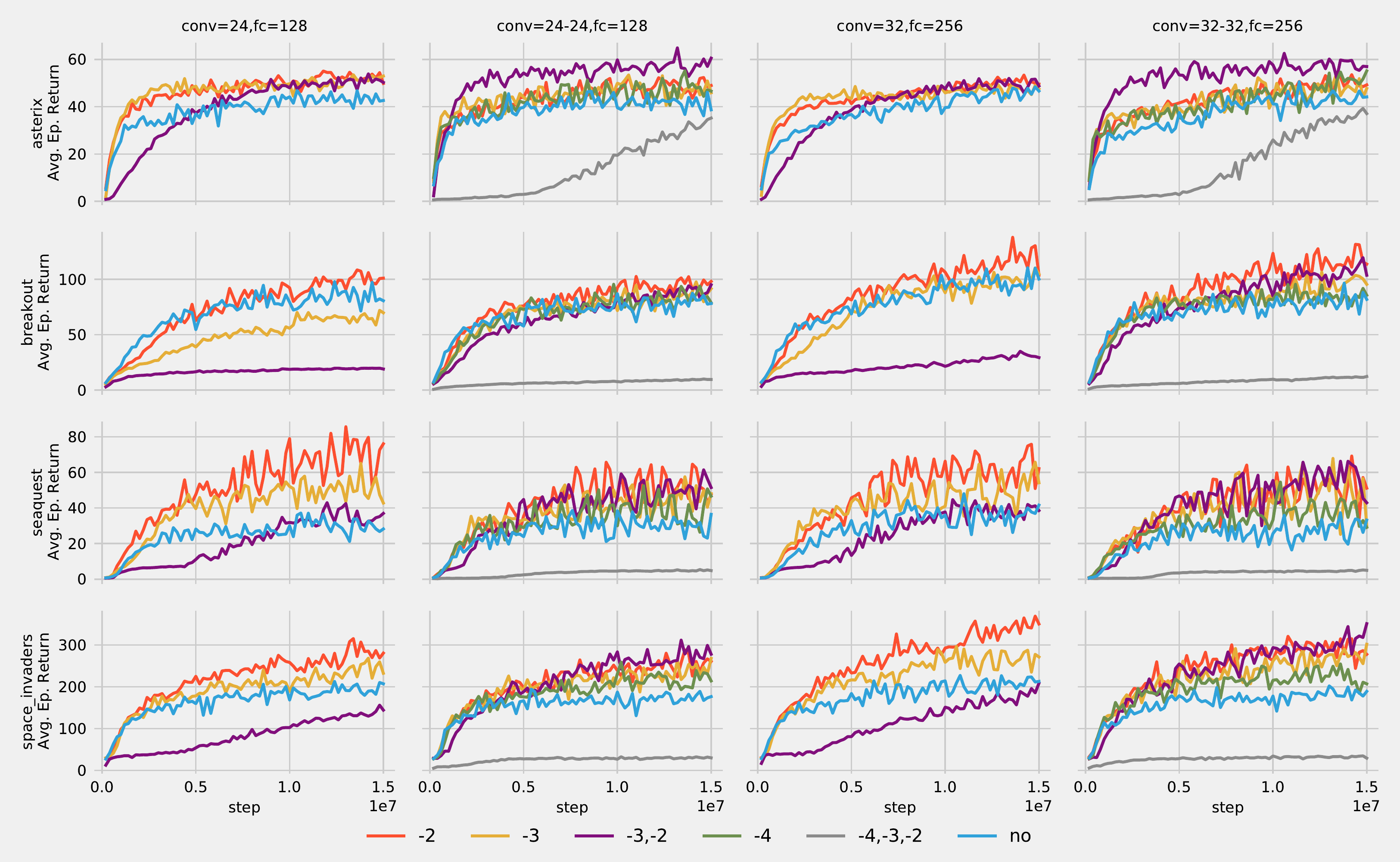}
    \caption{Performance curves of DQN agents using four different architectures trained for 15M steps. This plot shows that baseline plateaus, while spectrally normalised variants generally don't. See Fig.~\ref{app-fig:dqn-minatar-all-radii-2} for plots of the spectral radii for the same experiments.}
    \label{app-fig:dqn-15M-4arch}
\end{figure*}

\subsection{Adaptability to changing dynamics}
\label{sub:app-minatar--adaptability}

We noticed that in most experiments on MinAtar the DQN agent reaches its peak performance within the standard 5M steps and then it plateaus. Most of these training curves end in flat performance regimes. In contrast, when \ac{sn} is used on a single hidden layer, not only that agents surpass the baseline performance, but they also show continuous improvement with no signs of plateauing. We therefore asked what happens if we extend the training period and trained agents for 15M steps. As anticipated, our \emph{long} training experiments show that \ac{sn} agents show steep learning curves even after large numbers of steps (Fig.~\ref{app-fig:dqn-15M-4arch}), supporting our claim that \ac{sn} yields a better adapting optimiser.

\subsection{Spectral Schedulers}
\label{sub:app-ss}

\setcounter{figure}{12} 
\begin{figure*}[ht]
    \centering
    \includegraphics[width=\textwidth]{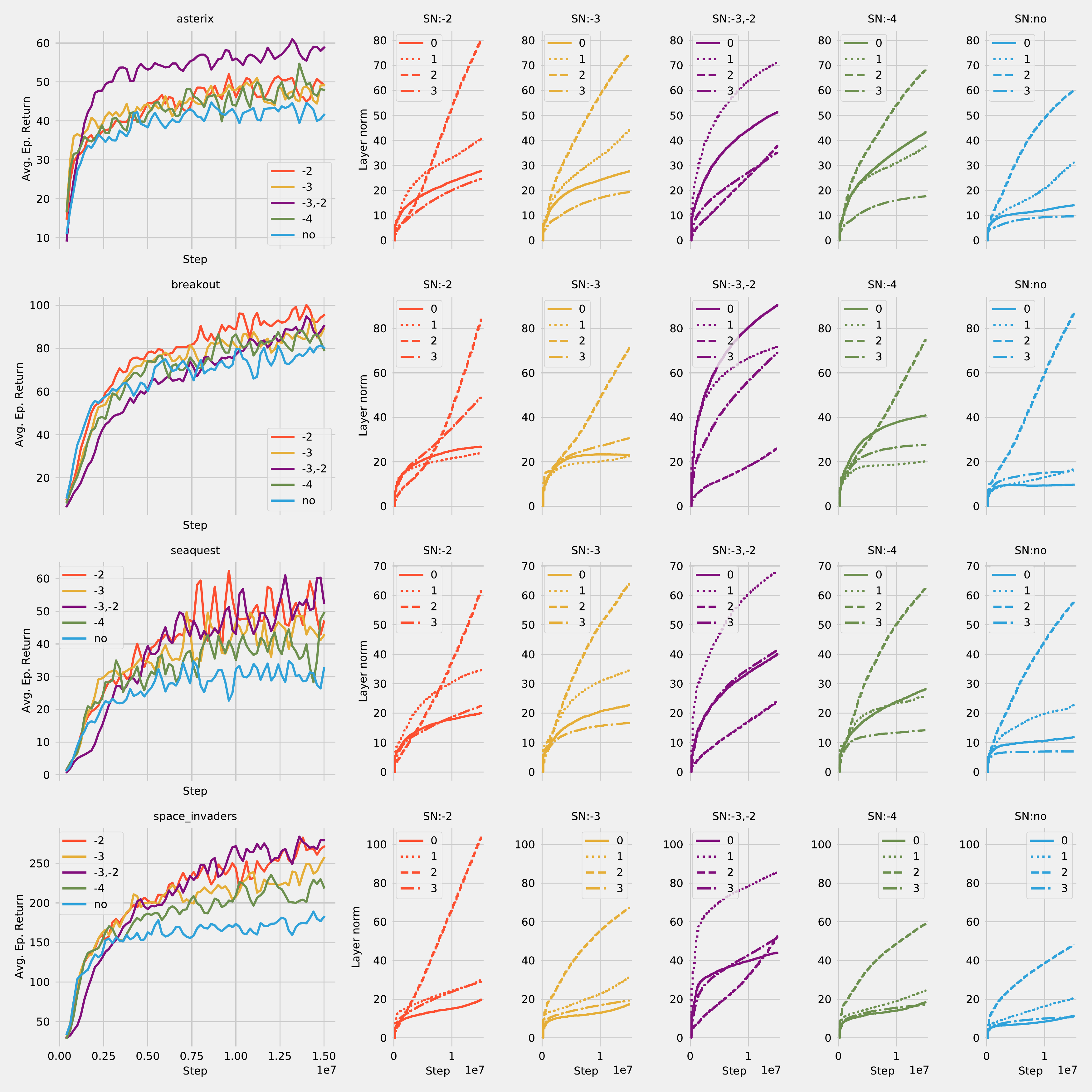}
    \caption{All spectral radii for the 15M experiment on MinAtar using a 4-layer architecture (conv=24-24,fc=128). Colors code the subsets of layers that are normalised (consistent with the rest of the document), while line styles code the four layers. Note how the penultimate layer has the largest spectral norm across all normalisation variants. 10 seeds.}
    \label{app-fig:dqn-minatar-all-radii-2}
\end{figure*}

For the experiments with the schedulers proposed in Sec.~\ref{sec:analysis--optimisation} we used the four estimator architectures in Table~\ref{tbl:app-ALL-models}. We detail the MinAtar Normalised Score for various subsets of layers considered in the comparison in Figure~\ref{app-fig:schedulers_detailed}. In a single subplot the lines represent the performance of the baseline, \ac{sn}, and spectral schedulers, all using the same spectral radii. Observe that \textsc{divOut} has a close behaviour to that of \ac{sn} not only on average, but also on a case base. In contrast, the \textsc{mulEps} and \textsc{divGrad} optimisers converge even when all hidden layers are normalised.
\setcounter{figure}{17}
\begin{figure*}[ht]
    \centering
    \includegraphics[width=\textwidth]{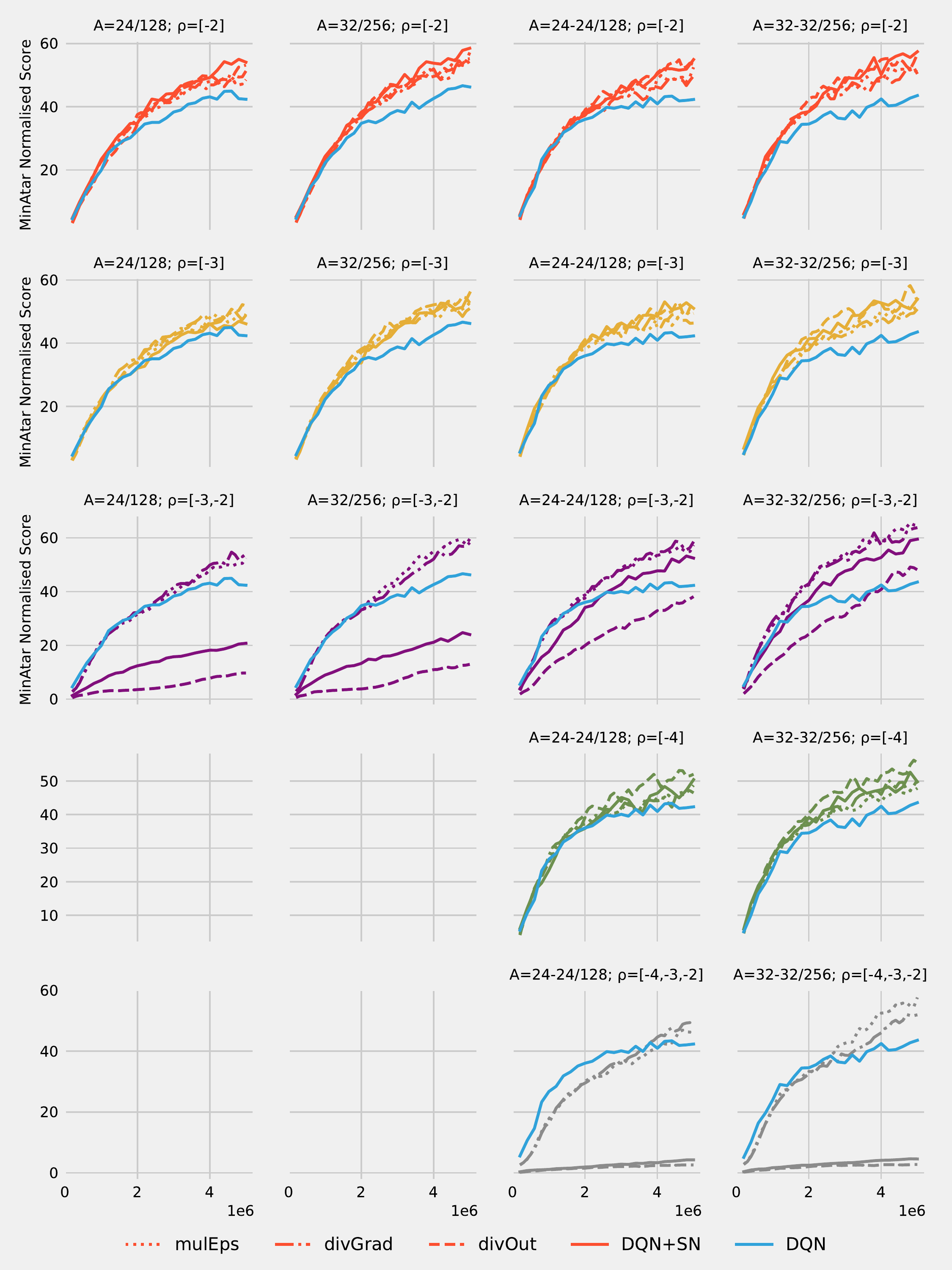}
    \caption{MinAtar Normalised Scored for the four architectures in Table~\ref{tbl:app-ALL-models} and various subsets of layers whose spectral radii are used for \ac{sn} or spectral schedulers. Notice that \textsc{divOut} behaves similarly to \ac{sn} (even when they fail to train), while \textsc{mulEps} and \textsc{divGrad} converge even when all hidden layers are normalised.}
    \label{app-fig:schedulers_detailed}
\end{figure*}

\begin{table}[h]
    \vskip 0.15in
    \begin{center}
        \begin{small}
            \begin{tabular}{lccc}
                \toprule
                Experiment & \thead{No of \\ conv layers}   & \thead{Conv \\ width}    & \thead{FC \\ width} \\
                \midrule
                \rule{0pt}{2ex}%
                \makecell[l]{Large optimisation \\ sweep (\ref{sub:app-minatar--optimisation_sweep})} &
                \makecell{
                   \textsc{1} \\ \textsc{2} \\ \textsc{3} \\
                   \textsc{4} \\ \textsc{2} \\ \textsc{3} \\
                } &
                \makecell{
                   \textsc{24} \\ \textsc{24} \\ \textsc{24} \\
                   \textsc{24} \\ \textsc{32} \\ \textsc{32} \\
                } &
                \makecell{
                   \textsc{128} \\ \textsc{128} \\ \textsc{128} \\
                   \textsc{128} \\ \textsc{256} \\ \textsc{256} \\
                }\\
                \midrule
                \rule{0pt}{2ex}%
                \makecell[l]{Regularisation (\ref{sub:app-minatar--regularisation})} &
                \makecell{
                   \textsc{1} \\ \textsc{3} \\ \textsc{1} \\
                   \textsc{3} \\ \textsc{1} \\ \textsc{3} \\
                } &
                \makecell{
                   \textsc{16} \\ \textsc{16} \\ \textsc{24} \\
                   \textsc{24} \\ \textsc{32} \\ \textsc{32} \\
                } &
                \makecell{
                   \textsc{64} \\ \textsc{64} \\ \textsc{128} \\ 
                   \textsc{128} \\ \textsc{256} \\ \textsc{256} \\ 
                }\\
                \midrule
                \rule{0pt}{2ex}%
                \makecell[l]{
                    Spectral schedulers  (\ref{sub:app-ss})\\
                    and long training run (\ref{sub:app-minatar--adaptability})
                } &
                \makecell{
                   \textsc{1} \\ \textsc{2} \\ \textsc{1} \\ \textsc{2} \\
                } &
                \makecell{
                   \textsc{24} \\ \textsc{24} \\ \textsc{32} \\ \textsc{32} \\
                } &
                \makecell{
                   \textsc{128} \\ \textsc{128} \\ \textsc{256} \\ \textsc{256} \\ 
                }\\
                \bottomrule
            \end{tabular}
        \end{small}
    \end{center}
    \caption{
        Architecture sets used in the experiments described in this section.
    }
    \label{tbl:app-ALL-models}
\end{table}

\clearpage

\section{Atari experiments}
\label{sec:app-atari}

\setcounter{figure}{20} 
\begin{figure*}[h!]
    \centering
    \includegraphics[width=\textwidth]{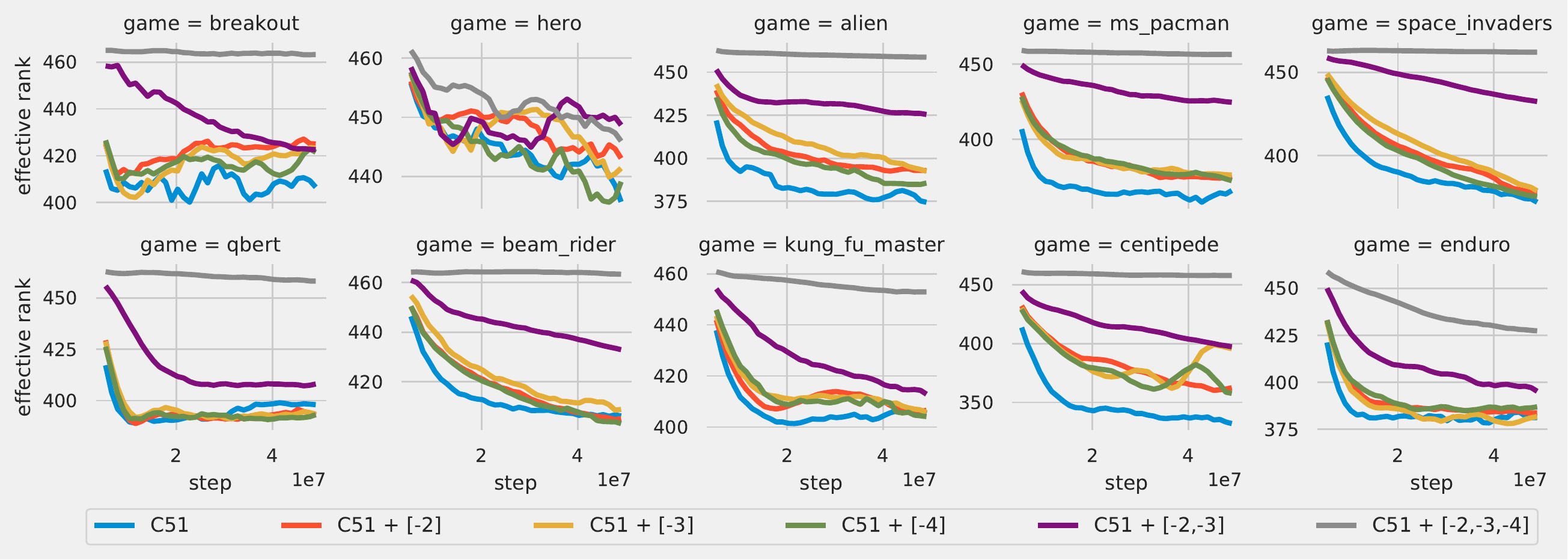}
    \caption{
        \textbf{Spectral Normalisation preserves the Effective Rank of the features.}
        Evolution of the effective rank of the features before the last linear layer of an C51 agent trained on 10 Atari games.
    }
    \label{app-fig:rank-atari}
\end{figure*}

\subsection{Evaluation protocols on Atari}
\label{sec:app-atari--evaluation}

In our work we mostly compare our \acf{ale} results with the \rainbow{} agent, therefore we adopt the evaluation protocol from  \cite{hessel2018RainbowCI}.
Every 250K training steps in the environment we suspend the learning and evaluate the agent on 125K steps (or 500K frames). All the agents we train on \ac{ale} follow this validation protocol, the only difference being the validation epsilon value: $\epsilon=0.001$ for C51 and DQN-Adam which we directly compare to \rainbow{} and uses the same value and $\epsilon=0.05$ for DQN-RMSProp which follows the exact same hyper-parameters from \cite{mnih2015human}.

A major difference between the \rainbow{} protocol and the \emph{null op starts} protocol used in earlier works is that in previous works the agent is evaluated for 30 or 100 \cite{van2016deep} episodes and is allowed to play up to $18,000$ frames (5 minutes of emulator time) or by the end of the episode, whatever came first, whereas we always evaluate for up to $500,000$ frames.

Episodes are limited at 108K steps, the agent receives a game over signal when losing a life as in previous works and we use the \emph{null op starts} to induce stochasticity in \ac{ale} games both at training and evaluation time \cite{van2016deep}.

\subsection{DQN-Adam}
\label{sec:app-atari--dqn_adam}

Next, we wanted to showcase \ac{sn} on an algorithm with a simpler objective such as \ac{dqn}. However our initial experiments on MinAtar suggested that \ac{sn} has a greater impact on Adam than on the RMSProp optimiser used in \ac{dqn} \cite{mnih2015human}. Since we also wanted to be able to compare our results with those of \rainbow{} we use similar hyper-parameters to those in \cite{hessel2018RainbowCI}. We list the full details in Table~\ref{tbl:app-dqn_adam_hyperparameters} especially since these hyper-parameters differ considerably from the the original DQN agent.

\begin{table}[h]
    \vskip 0.15in
    \begin{center}
        \begin{small}
            \begin{tabular}{lcc}
                \toprule
                \textsc{Hyper-parameter}        & \textsc{Value} \\
                \midrule
                \rule{0pt}{2ex}%
                discount $\gamma$               & \textsc{0.99} \\
                update frequency                & \textsc{4} \\
                target update frequency         & \textsc{8000} \\
                \midrule
                \rule{0pt}{1ex}%
                starting $\epsilon$             & \textsc{1.0} \\
                final $\epsilon$                & \textsc{0.01} \\
                $\epsilon$ steps                & \textsc{250000} \\
                $\epsilon$ schedule             & linear \\
                warmup steps                    & \textsc{20000} \\
                \midrule
                \rule{0pt}{1ex}%
                replay size                     & \textsc{1M} \\
                batch size                      & \textsc{32} \\
                history length                  & \textsc{4} \\
                \midrule
                \rule{0pt}{1ex}%
                cost function                   & \textsc{SmoothL1Loss} \\
                optimiser                       & \textsc{Adam} \\
                learning rate $\eta$            & \textsc{0.00025} \\
                damping term $\epsilon$         & \textsc{0.0003125} \\
                $\beta_1, \beta_2$              & \textsc{(0.9, 0.999)} \\
                \midrule
                \rule{0pt}{1ex}%
                validation steps                & \textsc{125000} \\
                validation $\epsilon$           & \textsc{0.001} \\
                \bottomrule
            \end{tabular}
        \end{small}
    \end{center}
    \caption{
        DQN-Adam hyper-parameters.
    }
    \label{tbl:app-dqn_adam_hyperparameters}
\end{table}

\subsection{DQN - RMSprop}
\label{sec:app-dqn-nature}

We also applied \ac{sn} to a DQN agent optimised with RMSprop as in the original \cite{mnih2015human}. In conjunction with RMSprop the impact of \ac{sn} seems minimal, far from the impressive improvement observed for the DQN-Adam agent. We leave for future work explaining the interaction between normalisation and RMSProp. See Table~\ref{tbl:hns-dqn-rmsprop} for comparing the Human Normalised Score of DQN-RMSprop to other agents, and Fig.~\ref{app-fig:dqn-rmsprop_ale_full} for individual plots per game.

\begin{table}[h]
    \vskip 0.15in
    \begin{center}
        \begin{small}
            \begin{tabular}{lcc}
                \toprule
                \textsc{Agent}                                  & \textsc{Mean}             & \textsc{Median} \\
                \midrule
                \rule{0pt}{2ex}%
                \textsc{DQN$^{\ast}$}         & \textsc{357.36}           & \textsc{102.94} \\
                \textsc{DQN SN[-2]} & \textsc{375.37}           & \textsc{105.19} \\
                \hline%
                \textsc{DQN} \cite{wang2016dueling}             & \textsc{216.84}           & \textsc{78.37}  \\
                \textsc{DQN-Adam$^{\ast}$}                      & \textsc{358.45}           & \textsc{119.45} \\
                \textsc{DQN-Adam SN[-2]}                        & \textsc{\textbf{719.95}}  & \textsc{\textbf{178.18}} \\
                \bottomrule
            \end{tabular}
        \end{small}
    \end{center}
    \caption{
        Mean and median Human Normalised Score on \dqngames{} Atari games with random starts evaluation.
        References indicate the sources for the scores for each algorithm.
        We mark our own implementations of the baseline with $\ast$. Our agents are evaluated with the protocol in \cite{hessel2018RainbowCI}. Note that the scores we report for our own implementation of DQN baseline are different from those reported in \cite{mnih2015human} because the evaluation protocol has changed.
    }
    \label{tbl:hns-dqn-rmsprop}
\end{table}

\subsection{Effective rank}
\label{sec:app-rank-atari}

Authors of \cite{kumar2020implicit} is making the case that for TD-learning with function approximation trained with SGD the neural network is being implicitly under-parametrised early in training.
Empirically they show this by looking at the \emph{effective rank} of the feature matrix $\bm{\Phi}$ which they approximate with the number of first $k$ singular values of $\bm{\Phi}$ that capture $99\%$ variance of all the singular values: $\sum_i^k \sigma_i(\bm{\Phi}) \;/\; \sum_j^d \sigma_j(\bm{\Phi}) \ge 0.99$.
In this case the feature matrix $\bm{\Phi}$ is the input to the last linear layer in our neural network.

We perform their experiment, this time with a \ac{c51} agent with and without normalised layers looking to better understand the regularisation effects of normalisation.
Figure~\ref{app-fig:rank-atari} shows an evolution of the effective rank for the baseline agent that is consistent with the report of \cite{kumar2020implicit}. Interestingly, the baseline agent is consistently the one making use of fewer and fewer dimensions in the feature space as training progresses while the normalised agents preserve the rank. We further corroborate this finding with that of \cite{miyato2018spectral} which is arguing that one of the possible disadvantages of \acf{wn} as opposed to \ac{sn} is that it prematurely producing sparse representations. %
Our experiments shows further shows that \ac{sn} helps with preserving the rank of the features early on in training even when compared to an un-normalised agent.

\clearpage
\setcounter{figure}{18} 
\begin{figure*}[ht]
    \centering
    \includegraphics[width=\textwidth]{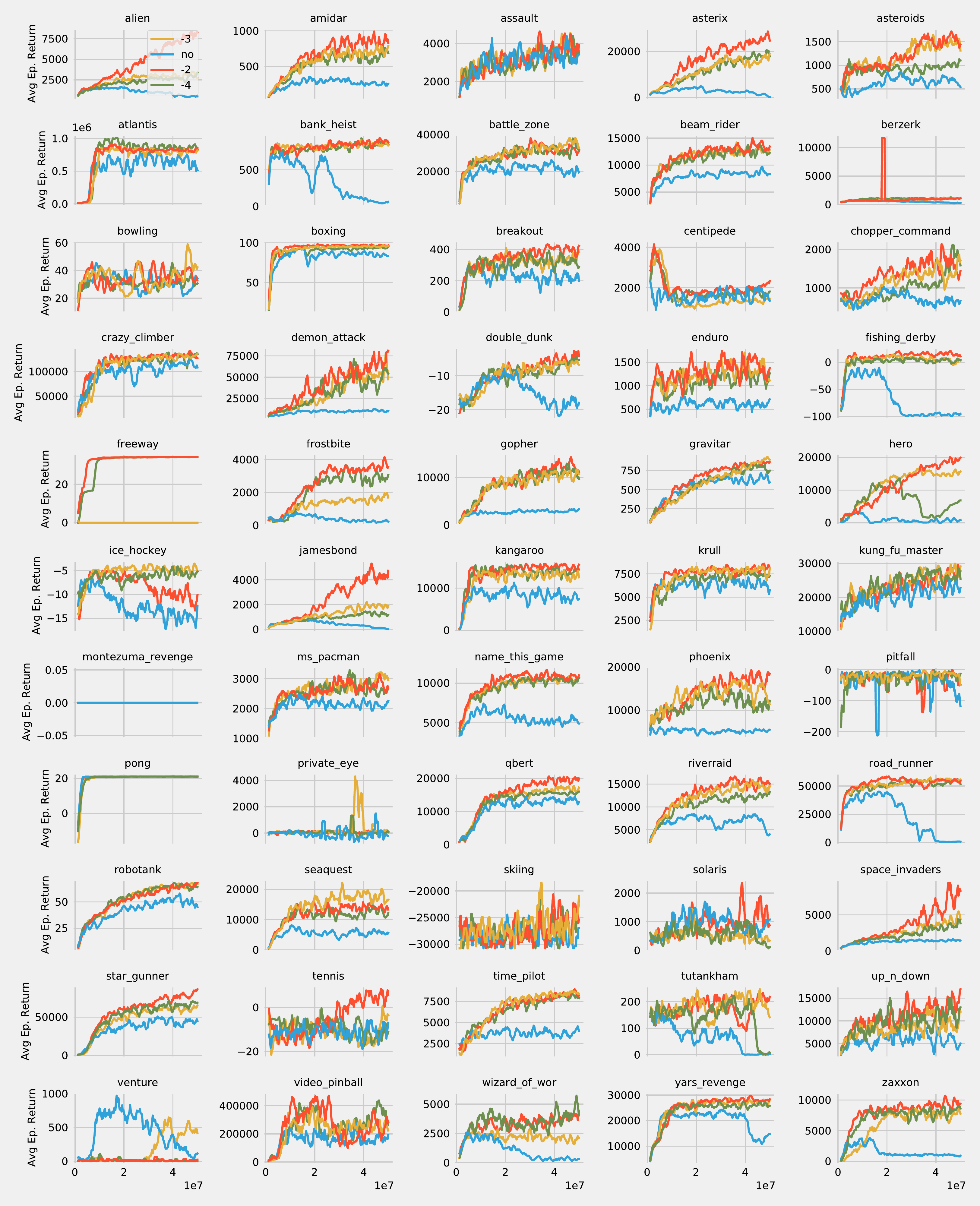}
    \caption{Performance curves of a DQN baseline optimised with Adam with \ac{sn} applied on three different layers.}
    \label{app-fig:dqn-adam_ale_full}
\end{figure*}

\begin{figure*}[ht]
    \centering
    \includegraphics[width=\textwidth]{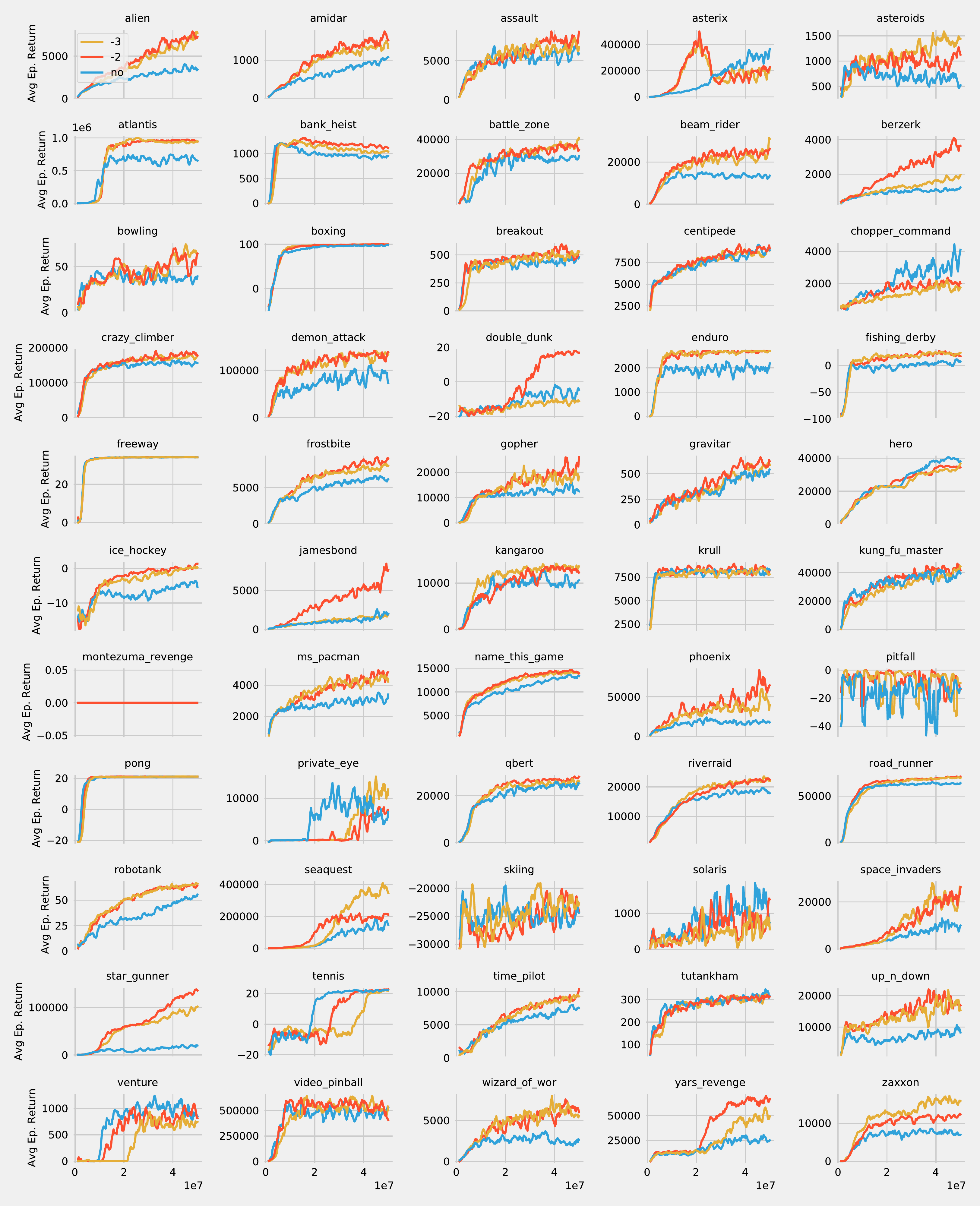}
    \caption{Performance curves of a C51 baseline with \ac{sn} applied on
        three different layers.}
    \label{app-fig:c51-icml_ale_full}
\end{figure*}
\setcounter{figure}{21}
\begin{figure*}[ht]
    \centering
    \includegraphics[width=\textwidth]{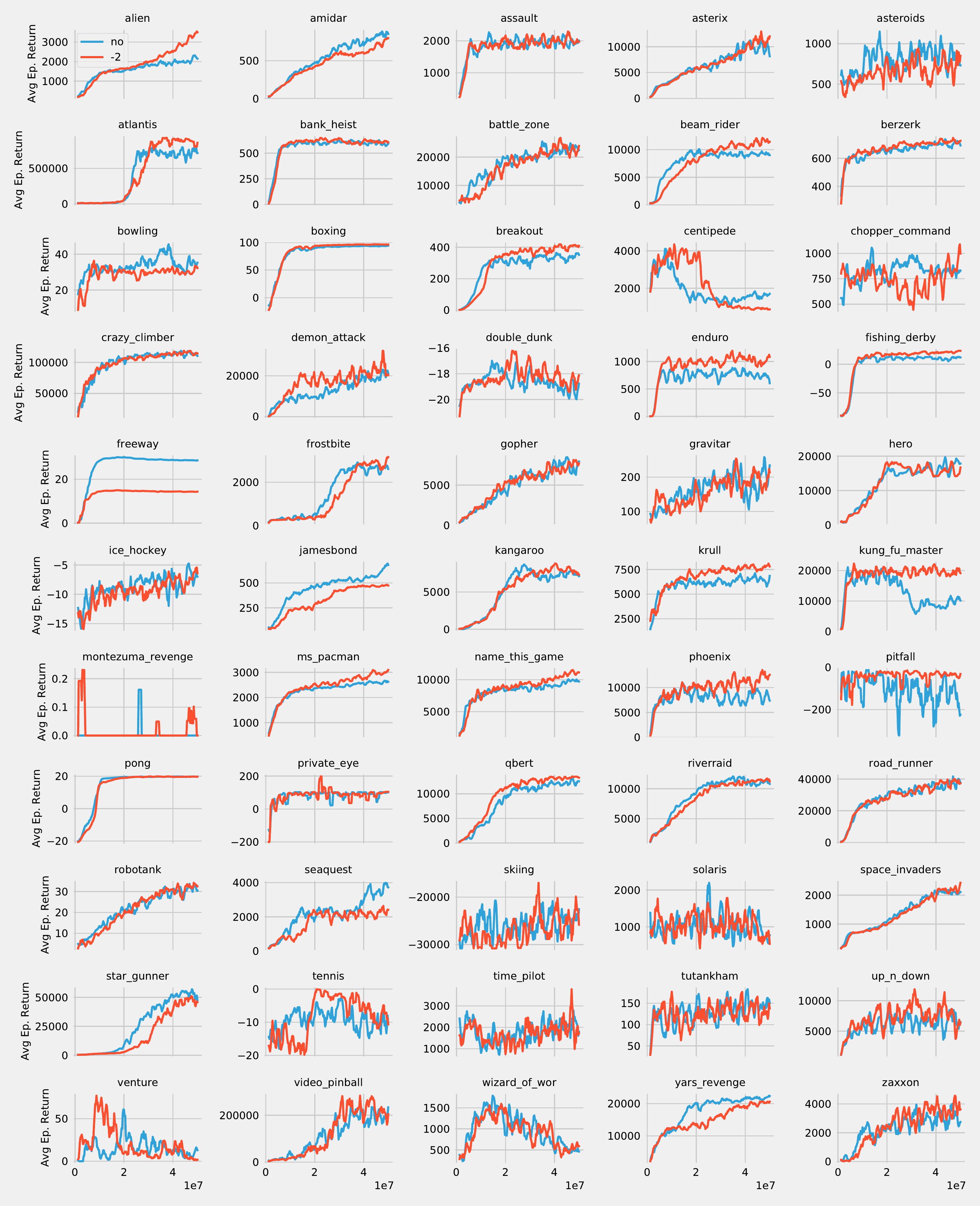}
    \caption{Performance curves of a DQN baseline optimised with RMSprop as in \cite{mnih2015human} with \ac{sn} applied on
        the penultimate layer.}
    \label{app-fig:dqn-rmsprop_ale_full}
\end{figure*}

\end{document}